\documentclass[11pt]{article}

\usepackage[preprint]{acl}

\usepackage{times}
\usepackage{latexsym} 

\usepackage[T1]{fontenc}
\usepackage[utf8]{inputenc}

\usepackage{microtype}

\usepackage{inconsolata}

\usepackage{graphicx}

\usepackage{amsmath, amsfonts}
\usepackage{tcolorbox}
\usepackage[table]{xcolor}

\definecolor{highlightbg}{RGB}{255, 240, 200}
\definecolor{red}{RGB}{220, 50, 50}
\definecolor{blue}{RGB}{0, 0, 205}

\usepackage{listings}
\usepackage{enumitem}
\usepackage{bm}
\usepackage{array}

\newcolumntype{M}[1]{>{\centering\arraybackslash}m{#1}}
\newcommand{\gc}{\cellcolor{gray!14}}

\usepackage{multirow}
\usepackage{booktabs}

\definecolor{entityBlue}{RGB}{0, 110, 180}
\definecolor{entityRed}{RGB}{180, 50, 50}
\definecolor{wrongParam}{RGB}{100, 100, 100}

\usepackage{tabularx}
\usepackage{colortbl}

\usepackage{mathtools}

\title{Do LLMs Know Tool Irrelevance? \\ Demystifying Structural Alignment Bias in Tool Invocations}

\author{
 \textbf{Yilong Liu\textsuperscript{1,2}},
 \textbf{Xixun Lin\textsuperscript{1,2}},
 \textbf{Pengfei Cao\textsuperscript{3}},
\\
 \textbf{Ge Zhang\textsuperscript{4}},
 \textbf{Fang Fang\textsuperscript{1,2}},
 \textbf{Yanan Cao\textsuperscript{1,2}}\thanks{Corresponding author.}
\\
 \textsuperscript{1}Institute of Information Engineering, Chinese Academy of Sciences, Beijing, China
\\
 \textsuperscript{2}School of Cyber Security, University of Chinese Academy of Sciences, Beijing, China
\\
 \textsuperscript{3}Institute of Automation, Chinese Academy of Sciences, Beijing, China
\\
 \textsuperscript{4}School of Computer Science and Technology, Donghua University, Shanghai, China
\\
  \texttt{\{liuyilong, caoyanan\}@iie.ac.cn}\\
}

\begin{document}
\maketitle
\begin{abstract}
Large language models (LLMs) have demonstrated impressive capabilities in utilizing external tools.
In practice, however, LLMs are often exposed to tools that are irrelevant to the user's query, in which case the desired behavior is to refrain from invocations. In this work, we identify a widespread yet overlooked mechanistic flaw in tool refusal, which we term structural alignment bias: Even when a tool fails to serve the user's goal, LLMs still tend to invoke it whenever query attributes can be validly assigned to tool parameters. To systematically study this bias, we introduce SABEval, a new dataset that decouples structural alignment from semantic relevance. 
Our analysis shows that structural alignment bias induces severe tool-invocation errors in LLMs, yet remains largely unaccounted for in existing evaluations. To investigate the internal mechanisms underlying this bias, we propose Contrastive Attention Attribution, which reveals two competing pathways for semantic checking and structural matching. The relative strength of these pathways drives LLMs' tool invocation decisions. Based on these findings, we further introduce a rebalancing strategy that effectively mitigates structural alignment bias, as demonstrated by extensive experiments, without degrading general tool-use capabilities. Our code and dataset are available at \url{https://github.com/along-l/irrelevant-tool}.
\end{abstract}

\section{Introduction}

\begin{figure*}[!h]
  \centering
  \includegraphics[width=0.875\linewidth,trim=0cm 7.325cm 0.70cm 0cm, clip]{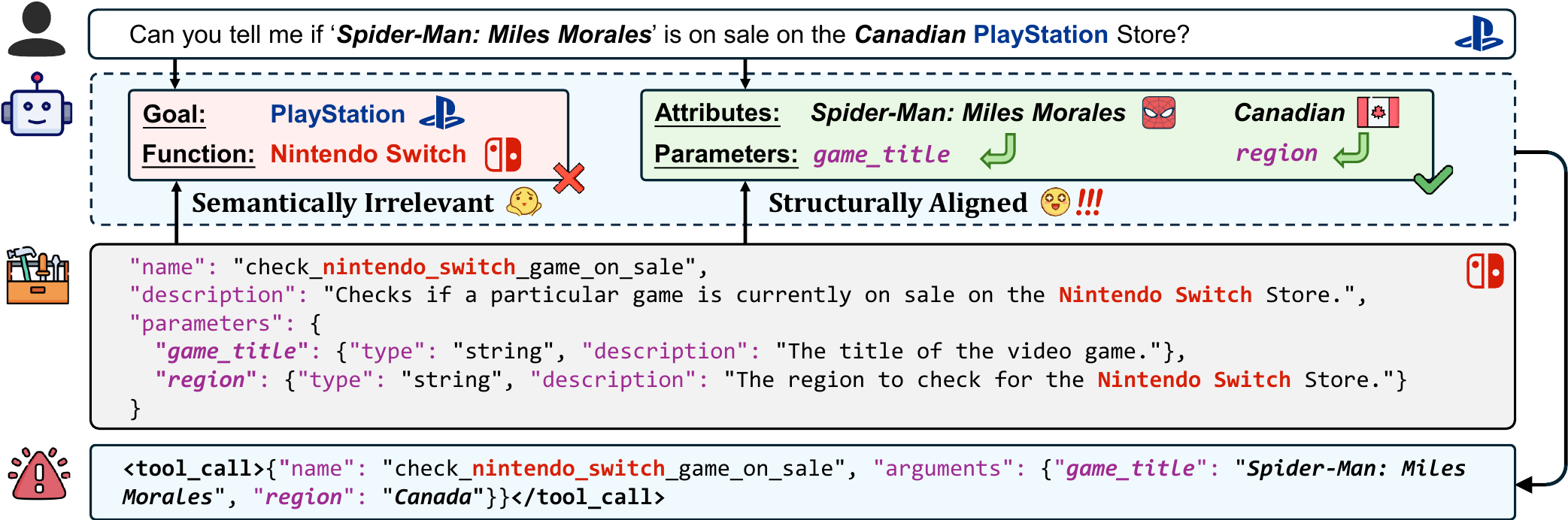}
  \caption{Illustration of erroneous tool invocation driven by the model's strong reliance on structural alignment.}
  \label{fig:main_demo}
\end{figure*} 

Driven by the significant breakthroughs in model architectures, computational capabilities, and large-scale data sources, large language models (LLMs) have advanced rapidly in recent years~\cite{naveed2025comprehensive,xi2025rise, mohammadi2025evaluation}. A key capability of LLMs is the utilization of external tools, which bridges the gap between text generation and real-world environmental interaction~\cite{gao2024confucius,houliston2025provable,qu2025tool}.
\par
However, in many practical scenarios, LLMs are often exposed to irrelevant tools that are useless for addressing user queries~\cite{zhang-etal-2024-toolbehonest, ross-etal-2025-when2call, liu2025wtu, lin2025llm}.
% For example, a user may ask for weather information, but the LLM only has access to a calculator tool.
For example, a user may request to refund a transaction, but the LLM only has access to a tool for transferring funds.
Invoking such irrelevant tools may introduce system latency and misleading outputs, and 
more critically, for high-risk applications, such as financial transactions or database management, can lead to irreversible and catastrophic consequences, including financial loss or data corruption~\cite{xia2025safetoolbench, subramani-etal-2025-mice}.
As a result, the desired behavior of LLMs is to recognize the limitation of the available tools and refrain from invocations.

Although recent studies have explored the ability of LLMs to reject irrelevant tools~\cite{xu2024reducing,patil2025the}, we find that there exists a widespread yet overlooked mechanistic flaw in LLMs when refusing tool invocations,
which we term \textbf{Structural Alignment Bias}.
As illustrated in Fig.~\ref{fig:main_demo}, while a tool cannot help realize the user's goal (e.g., user queries a game on \textit{PlayStation}, but the provided tool is designed to check \textit{Nintendo Switch} games), the model still shows a strong tendency to invoke this tool as there exists a valid one-to-one assignment between query attributes and tool parameters (e.g., the attribute ``Spider-Man'' can be validly assigned to ``\texttt{game\_title}'', and ``Canadian'' to ``\texttt{region}''). 
\par
In this paper, we define the inconsistency between the tool's function and the user's goal as \textbf{Semantic Irrelevance}, 
and refer to the existence of such one-to-one assignment as \textbf{Structural Alignment}. Building upon these definitions, we characterize structural alignment bias as a systematic tendency of LLMs to prioritize this structural alignment as a shortcut for tool invocations, thereby bypassing the verification of semantic relevance.

To further verify and analyze the proposed structural alignment bias, we construct \textbf{SABEval}, a dataset designed to decouple structural alignment from semantic relevance.
SABEval mirrors real-world scenarios where distinct services share unified interfaces, avoiding synthetic corner cases.
We conduct comprehensive experiments on five widely used tool-augmented LLMs.
Our behavioral analysis uncovers a striking contrast: while the error rate is negligible ($<0.2\%$) on structurally misaligned samples created by previous studies (See \S~\ref{sec:formulation} for details.), it rises sharply to 41.9\% under structural alignment and further escalates to 90.4\% as the alignment degree increases. Furthermore, we perform the counterfactual analysis which confirms a strong causal relationship between structural alignment and erroneous invocations, highlighting the critical impact of structural alignment bias on model behaviors.

To investigate the internal mechanisms underlying structural alignment bias, we introduce \textbf{Contrastive Attention Attribution} (CAA), a novel interpretability method that traces the information flow without requiring the strict token-wise alignment between counterfactual pairs. Our method reveals two distinct pathways responsible for semantic checking and structural matching, respectively. Moreover, we find that LLMs' tool invocation decisions are driven by the relative strength of these two pathways. Based on these insights, we propose a new strategy that rebalances the two pathways to mitigate structural alignment bias. Extensive experiments demonstrate that our method can significantly alleviate structural alignment bias across diverse models and tools, while maintaining their general tool-use capabilities. Our main contributions are summarized as follows:
\begin{itemize}[leftmargin=*,itemsep=0pt, topsep=0pt, parsep=0pt]
    \item \textbf{Problem Identification.} We are the first to identify and formalize structural alignment bias, a mechanistic flaw in which models over-rely on structural alignment when making tool invocation decisions.
    
    \item \textbf{Empirical Evidence.} We construct SABEval, a dataset that decouples structural alignment from semantic relevance and provides comprehensive evidence of the prevalence, severity, and causality of structural alignment bias.
    
    \item \textbf{Interpretability Method.} We propose CAA, a novel interpretability method that reveals models' internal computations without relying on the strict token correspondence of counterfactuals, enabling the discovery of competitive information flow for tool invocations.
    
    \item \textbf{Bias Mitigation.} We introduce a precise intervention to mitigate structural alignment bias. Extensive experiments show its effectiveness, achieving an 80\% relative error reduction while preserving general tool-use capabilities.
\end{itemize}

\section{Problem Formulation}
\label{sec:formulation}
In this section, we first introduce the irrelevance setting in tool invocations. We then propose a formal concept of structural alignment with our core question.

\paragraph{Tool Invocations with Irrelevant Tools.}
A critical challenge in tool-augmented LLMs is whether LLMs can correctly refuse to invoke tools that cannot serve the user's goal, a scenario often referred to as the \textbf{Irrelevance Setting}~\cite{patil2025the}. Our work specifically addresses this problem. Consider an input $(t, q)$, where $t$ is a candidate tool and $q$ is the user query. We denote the tool's function by $f_t$ and the user's goal by $g_q$. In the irrelevance setting, $t$ is \emph{semantically irrelevant} to $q$, which we formalize as $f_t \not\equiv g_q$. This implies that $f_t$ does not help realize $g_q$, and the model should reject $t$~\cite{chen2025acebench}. For example, as illustrated in Fig.~\ref{fig:main_demo}, the tool is designed to \textit{check Nintendo Switch games} (i.e., $f_t$), which fails to address the user's goal of \textit{checking PlayStation games} (i.e., $g_q$), and  LLMs should refuse this invocation. 

\paragraph{Structural Alignment.}
Besides semantic relevance, we hypothesize that an LLM's decision to invoke tools is strongly influenced by another factor: whether the attributes expressed in $q$ can be validly mapped to the parameters of $t$. We refer to this factor as structural alignment as described in the Introduction.
\par
Formally, let $\mathcal{P}_t = \{p_1, \dots, p_n\}$ be the parameters of $t$ (e.g., ``\texttt{game\_title}'', ``\texttt{region}''),
and $\mathcal{A}_q = \{a_1, \dots, a_m\}$ be the attributes extracted from $q$ (e.g., ``Spider-Man'', ``Canadian''). We introduce a binary indicator:
\begin{equation}
    b_{t,q}
    = \mathbb{I}\big( \mathcal{A}_q \cong \mathcal{P}_t \big),
\end{equation}
where $\mathbb{I}(\cdot)$ is the indicator function and $\mathcal{A}_q \cong \mathcal{P}_t$ denotes that there exists a \emph{bijection}\footnote{That is, $\mathcal{A}_q$ and $\mathcal{P}_t$ are in a one-to-one correspondence, implying $|\mathcal{A}_q| = |\mathcal{P}_t|$.}
between $\mathcal{A}_q$ and $\mathcal{P}_t$ such that every $a_j \in \mathcal{A}_q$ can be validly assigned to a parameter $p_i \in \mathcal{P}_t$, and vice versa (e.g., ``Spider-Man'' for ``\texttt{game\_title}'' and ``Canadian'' for ``\texttt{region}'' in Fig.~\ref{fig:main_demo}, yielding $b_{t,q}=1$). For a given input $(t, q)$, we suppose that structural alignment holds when $b_{t,q}=1$, and that structural misalignment exists when $b_{t,q}=0$.
\par
The phenomenon of structural alignment has been largely ignored by previous evaluations of models' ability to reject irrelevant tools~\cite{patil2025the, chen2025acebench, huang2023metatool,liu2025toolace}. They typically construct evaluation samples by randomly pairing a query with tools sampled from a pool. While such constructions ensure semantic irrelevance, they predominantly induce structural misalignment, thereby confounding evaluation results. This observation raises the core question of our work: \textit{Do LLMs genuinely recognize that semantic relevance is required for a valid tool invocation, or do they merely rely on structural alignment to make decisions?}

\section{Experimental Setup}
\label{sec:dataset}
To answer the above question, we first establish the experimental setup for our empirical study. Specifically, we construct a new dataset, SABEval, and describe its extensions, as well as the evaluated models used in our experiments.

\begin{figure}[!t]
  \centering
  \includegraphics[width=1\linewidth,trim=0cm 14.75cm 6.8cm 0cm, clip]{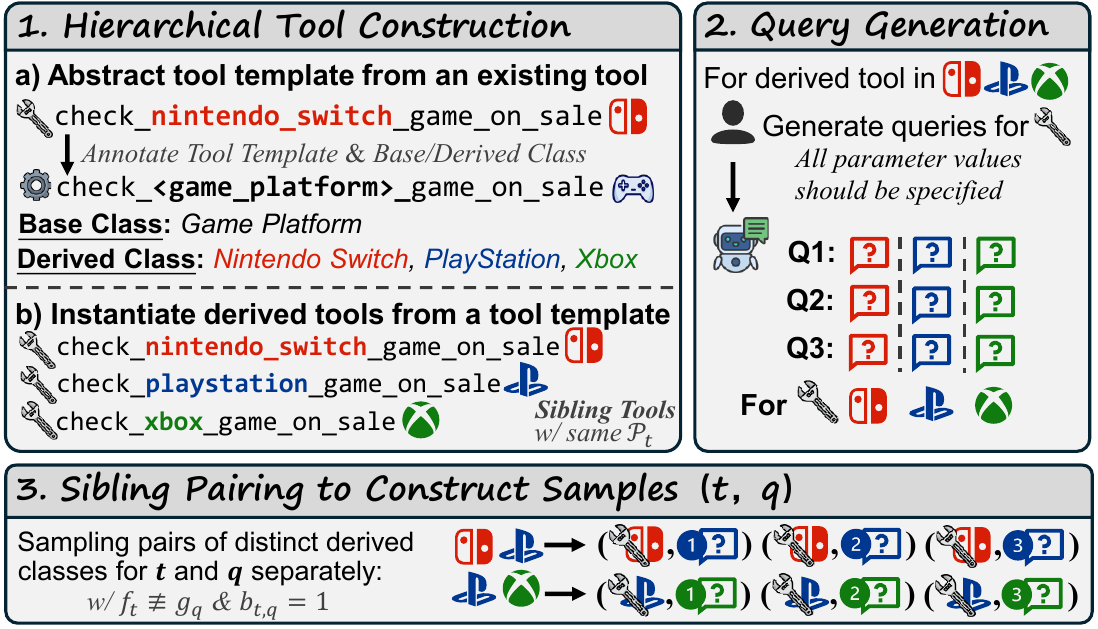}
  \caption{SABEval construction with three steps. Samples are semantically irrelevant but structurally aligned.}
  \label{fig:dataset_demo}
\end{figure}

\paragraph{Dataset.}
Existing datasets entangle semantic irrelevance with structural misalignment.
To strictly isolate the former (i.e., $f_t\not\equiv g_q$ \& $b_{t,q}=1$), we construct a new dataset \textbf{SABEval},  based on the \textit{polymorphism} principle of Object-Oriented Programming~\cite{meyer1997object}.
As illustrated in Fig.~\ref{fig:dataset_demo}, the construction pipeline comprises three steps: hierarchical tool construction, query generation, and sibling pairing.
Crucially, no valid tool is available for any query in SABEval, meaning any tool invocation constitutes an error.
Construction details are provided in Appendix~\ref{app:dataset}.
This design mirrors real-world scenarios where distinct services share unified interfaces, rather than creating synthetic corner cases.
Finally, we construct \textit{101} tool templates, \textit{5} queries per derived tool, and \textit{10} sibling combinations per tool template. This process yields our base dataset $\mathcal{D}_0$, comprising \textit{5,050} instances.

\paragraph{Dataset Extensions.}
To support the analysis in \S\ref{sec:behavior} and \S\ref{sec:mech}, we construct four extensions $\mathcal{D}_1, \dots, \mathcal{D}_4$ by adding tool templates of $\mathcal{D}_0$ with $k\in\{1,2, 3,\allowbreak 4\}$ additional parameters\footnote{We employ GPT-4o~\cite{gpt4o} to synthesize additional parameters.}.
We follow the pipeline in Fig.~\ref{fig:dataset_demo} to generate derived tools and queries from new tool templates, while reusing the original sibling pairings of $\mathcal{D}_0$ to construct samples.
This guarantees that $\mathcal{D}_{k+1}$ differs from $\mathcal{D}_k$ by one additional attribute-parameter assignment.

\paragraph{Evaluated Models.}
We investigate five models that natively support tool invocations: the Qwen3 series (4B, 8B, and 14B)~\cite{yang2025qwen3}\footnote{For the Qwen3 series, we disable the ``thinking mode'' to maintain consistency with the other models. Nevertheless, we report results with thinking mode enabled in Appendix~\ref{app:thinking_mode}. We also evaluate Qwen3-32B to study the impact of model size in Appendix~\ref{app:model_size}.}, ToolACE-2.5-8B~\cite{liu2025toolace}, and Watt-Tool-8B~\cite{watt-tool-8B}.
ToolACE-2.5-8B and Watt-Tool-8B are fine-tuned from Llama-3.1-8B-Instruct~\cite{dubey2024llama}\footnote{We exclude the original Llama models due to their inability to consistently adhere to the tool invocation format.}. We adopt the default system prompts for all evaluated models, and more details are given in Appendix~\ref{sec:apd_models}.

\section{Behavior Analysis}
\label{sec:behavior}
To address the research question proposed in \S~\ref{sec:formulation}, we tackle the following two progressive inquiries:
\begin{enumerate}[leftmargin=*, itemsep=0pt, topsep=2pt, parsep=1pt]
    \item How do LLMs perform with structurally aligned but semantically irrelevant tools? (\S~\ref{subsec:phenomenon})
    \item To what extent does structural alignment contribute to erroneous tool invocations? (\S~\ref{sec:causal})
\end{enumerate}

\subsection{Performance under Structural Alignment}
\label{subsec:phenomenon}
This section evaluates the model's capability to reject semantically irrelevant but structurally aligned tools. We first introduce the used metric: \textbf{Tool Invocation Rate} (TIR). Consider a dataset $\mathcal{D}$ consisting of $N$ samples. Let $P(w|q,t)$ denote the model's predicted probability for the next token $w$. TIR is defined as the proportion of samples where the special tool-call token $w_{\text{tool}}$ is the most likely prediction:
\begin{equation}
    \text{TIR} = \frac{1}{N} \sum_{i=1}^{N} \mathbb{I}\left( \operatorname*{argmax}_{w \in \mathcal{V}} P(w|q,t) = w_{\text{tool}} \right),
\end{equation}
where $\mathcal{V}$ is the vocabulary. $w_\text{tool}$ for all models are shown in Appendix~\ref{app:special_tokens}.
This formulation enables efficient evaluation via a single forward pass without full response generation. While it theoretically counts all tool-call format outputs, including hallucinated tool invocations, our premise is that models faithfully invoke only the provided tools. We verify this in Appendix~\ref{app:special_tokens}. For models prone to tool hallucination, a fine-grained metric based on full response parsing would be necessary.
In our setting, a lower TIR indicates a stronger capability to reject irrelevant tools. We then report two key empirical observations.

\begin{table}[t]
  \centering
  \small
  \begin{tabular}{p{2.2cm}M{1.2cm}M{1.2cm}M{0.8cm}}
    \toprule
    \textbf{Model} & \textbf{Random} & \textbf{SABEval} & $\bm{\Delta}$ \\
    \midrule
    Qwen3-4B & 0.16 & \textbf{40.04} & +39.88\\
    Qwen3-8B & 0.04 & \textbf{34.26} & +34.22\\
    Qwen3-14B & 0.04 & \textbf{41.86} & +41.82\\
    ToolACE-2.5-8B & 0.12 & \textbf{40.67} & +40.55\\
    Watt-Tool-8B & 0.00 & \textbf{10.51} & +10.51\\
    \bottomrule
  \end{tabular}
  \caption{Tool invocation rate (\%) on randomly paired samples and SABEval ($\mathcal{D}_0$).}
  \label{tab:performance}
\end{table}

\paragraph{High Rates of Erroneous Invocations.}
We conduct experiments on $\mathcal{D}_0$ of SABEval and compare the results with the random pairing setting (uniformly sampling a tool for every query).
As shown in Table~\ref{tab:performance}, across all models, TIR of random pairing remains extremely low (max 0.16\%), while they surge dramatically (max 41.86\%) on SABEval. This substantial gap indicates that LLMs hardly reject irrelevant tools when they are structurally aligned, suggesting that their decisions could be strongly biased toward structural alignment.
More importantly, this gap indicates that the random pairing setting substantially overestimates LLMs' ability to reject irrelevant tools.

\begin{figure}[t]
  \centering
  \includegraphics[width=\columnwidth]{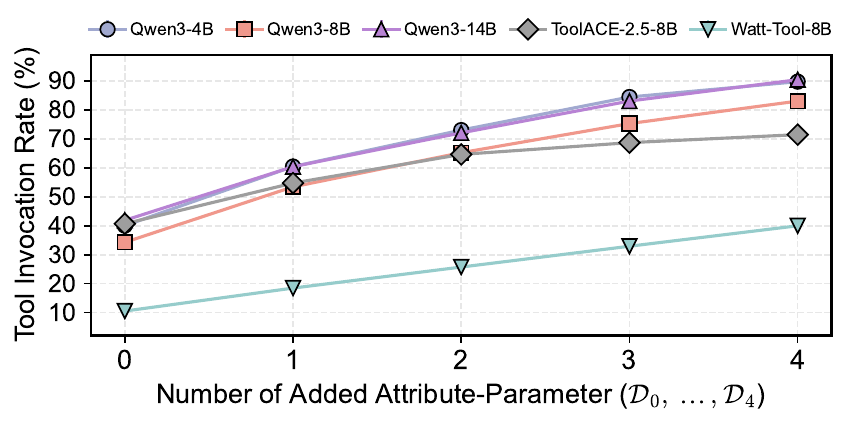}
  \caption{Tool invocation rate (\%) on SABEval subsets $\mathcal{D}_0,\dots ,\mathcal{D}_4$ with increasing structural alignment degree.}
  \label{fig:alignment_degree}
\end{figure}

\paragraph{Error Scales with Alignment Degree.}
Intuitively, more fillable tool parameters may make models more likely to invoke the tool.
We therefore examine whether TIR increases systematically with the number of attribute--parameter assignments between $q$ and $t$, which we term the \textbf{structural alignment degree}.
To this end, we evaluate on $\mathcal{D}_0,\dots,\mathcal{D}_4$ of SABEval\footnote{We do not simply group samples by the number of parameters in $\mathcal{D}_0$, as parameter counts are unevenly distributed across tool templates, and the template itself acts as a confounding factor between TIR and alignment degree.}.
As illustrated in Fig.~\ref{fig:alignment_degree}, we find the error rates of all models consistently and significantly increase as the alignment degree grows (up to 90.44\%). Results grouped by tool template show the same trend (See Appendix~\ref{app:group_tt} for details.). This further underscores the severity of the performance collapse on structurally aligned samples and reveals the fundamental flaw in the models.

\begin{table*}[t]
\centering
\small
\begin{tabularx}{\textwidth}{@{} l l >{\raggedright\arraybackslash}X >{\raggedright\arraybackslash}X @{}}
\toprule

\multicolumn{4}{l}{\cellcolor{gray!10}\textbf{User Query:} Can you tell me if `\textcolor{entityBlue}{Spider-Man: Miles Morales}' is on sale on the \textcolor{entityRed}{Canadian} \textbf{PlayStation} Store?} \\
\multicolumn{4}{l}{\cellcolor{gray!10}\textbf{Original Tool:} \texttt{check\_\textbf{nintendo\_switch}\_game\_on\_sale} \quad \textbf{Params:} \texttt{\textcolor{entityBlue}{game\_title}}, \texttt{\textcolor{entityRed}{region}}} \\
\midrule

\textbf{Counterfactual} & \textbf{Strategy} & \textbf{Tool Name} & \textbf{Parameters} \\
\midrule

\multirow{3}{*}{Structural} 
& \textit{Param Substitution} 
& \texttt{check\_\textbf{nintendo\_switch}\_game\_on\_sale}
& \texttt{\textcolor{entityBlue}{game\_title}}, \texttt{\textcolor{wrongParam}{include\_dlc}} \\ 

\cmidrule(r){2-4}

& \textit{Param Removal} 
& \texttt{check\_\textbf{nintendo\_switch}\_game\_on\_sale}
& \texttt{\textcolor{entityBlue}{game\_title}} \\

\cmidrule(r){2-4}

& \textit{Param Addition} 
& \texttt{check\_\textbf{nintendo\_switch}\_game\_on\_sale}
& \texttt{\textcolor{entityBlue}{game\_title}}, \texttt{\textcolor{entityRed}{region}}, \texttt{\textcolor{wrongParam}{include\_dlc}} \\

\midrule

Semantic
& \textit{Target Tool}
& \texttt{check\_\textbf{playstation}\_game\_on\_sale}
& \texttt{\textcolor{entityBlue}{game\_title}}, \texttt{\textcolor{entityRed}{region}} \\

\bottomrule

\end{tabularx}
\caption{Example of counterfactuals. We construct \textbf{structural counterfactuals} by intervening on the original tool parameters via three strategies (\textit{substitution}, \textit{removal}, and \textit{addition}) while retaining the original query. The \textbf{semantic counterfactual} includes a valid tool that serves the user's goal, which is utilized for pathway discovery in \S~\ref{sec:method}.}
\label{tab:intervention_examples}
\end{table*}

\subsection{Causal Role of Structural Alignment}
\label{sec:causal}
In this section, we answer the question of to what extent structural alignment contributes to the erroneous invocations observed in \S~\ref{subsec:phenomenon}.
To do so, we analyze structural counterfactuals of SABEval by intervening on the structural alignment status.

\paragraph{Intervention Design.}
Given a sample $(t,q)$ with $\mathcal{P}_t=\{p_{1}, \dots, p_n\}$ and $b_{t,q}=1$, we intervene on $t$, i.e., $t \rightarrow t^*$, and pair it with $q$ to construct the counterfactual $(t^*,q)$, for which $b_{t^*,q}=0$.
Specifically, we implement three intervention strategies, as illustrated in Table~\ref{tab:intervention_examples}:
\begin{itemize}[leftmargin=*,itemsep=0pt, topsep=0pt, parsep=0pt]
    \item \textbf{Parameter Substitution.} We substitute $l$ original parameters, such that $\mathcal{P}_{t^*} = \{p_1, \dots, p_{n-l},\allowbreak p^*_1, \dots, p^*_l\}$, where the new parameters $p^*_i$ are mutually distinct and do not appear in $\mathcal{P}_t$. In this case, there exist both unassignable user attributes and unspecified parameters.
    \item \textbf{Parameter Removal.} We remove $l$ original parameters, such that $\mathcal{P}_{t^*} = \{p_1, \dots, p_{n-l}\}$. In this case, $|\mathcal{A}_q| > |\mathcal{P}_{t^*}|$, representing a scenario where user attributes cannot be fully assigned.
    \item \textbf{Parameter Addition.} We append $l$ new parameters, such that $\mathcal{P}_{t^*} = \{p_1, \dots, p_n, p^*_1, \dots,\allowbreak p^*_l\}$. In this case, $|\mathcal{A}_q| < |\mathcal{P}_{t^*}|$, representing a scenario where some parameters are unspecified.
\end{itemize}
In practice, we set the perturbation size $l=1$. We conduct experiments on the $\mathcal{D}_1$ subset of SABEval and utilize the additional parameters generated in \S~\ref{sec:dataset}, thereby ensuring the plausibility of the counterfactual tools.

\begin{figure}[t]
  \centering
  \includegraphics[width=\columnwidth]{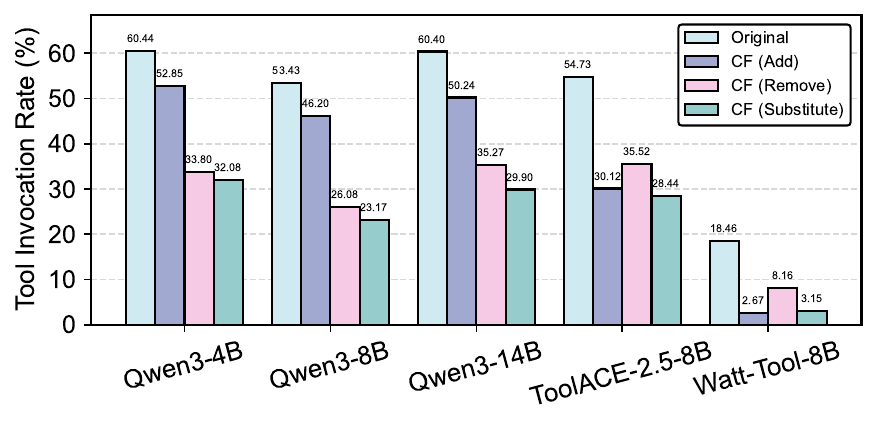}
  \caption{Tool invocation rate (\%) on original samples and their structural counterfactuals from $\mathcal{D}_1$.}
  \label{fig:tir_cf}
\end{figure}

\paragraph{Empirical Results.}
As shown in Fig.~\ref{fig:tir_cf}, all intervention strategies lead to a pronounced reduction in TIR. Notably, parameter substitution exhibits the most significant and consistent impact across all models (relative declines of 58\% for Qwen3-8B and 83\% for Watt-Tool), aligning with its effect of introducing unassignable attributes and unspecified parameters. 
We further investigate the probability shifts of $w_{\text{tool}}$ in Appendix~\ref{app:prob_shift}, where over 70\% and 90\% of samples exhibit clear probability decreases for the Qwen3 family and other models, respectively. Overall, these results underscore a strong causal effect of structural alignment on erroneous tool invocations.

\paragraph{Structural Alignment Bias.} 
We previously show that LLMs are prone to invoking irrelevant tools under structural alignment (\S~\ref{subsec:phenomenon}). Here, we further establish a strong causal link between such behavior and structural alignment. Based on these analyses, we can now answer the core question posed in \S~\ref{sec:formulation}: \textit{Instead of genuinely recognizing the necessity of semantic relevance, LLMs exhibit a systematic bias toward structural alignment in tool invocations.} We term this important problem structural alignment bias\footnote{We provide further analysis in Appendix~\ref{app:text_sim} to distinguish this bias from naive textual similarity, confirming it originates from the structural properties of the data.}. Such bias prevents LLMs from reliably rejecting irrelevant tools, leading to erroneous tool invocations and potentially irreversible consequences.

\section{Mechanistic Analysis}
\label{sec:mech}
In this section, we investigate the internal mechanism driving structural alignment bias.
We consider semantic irrelevance and structural alignment as competing signals and propose a novel method CAA (\S~\ref{sec:method}), which reveals the specific information flow responsible for processing them: \textit{semantic pathways} (checking the semantic irrelevance, then suppressing invocations) and \textit{structural pathways} (matching attributes to parameters, then promoting invocations). Our analysis suggests that their relative strength drives models' decisions, and targeted interventions offer an effective avenue for mitigating this bias (\S~\ref{sec:mech_result}).

\subsection{Contrastive Attention Attribution}
\label{sec:method}
Attention heads serve as channels for information flow between positions~\cite{elhage2021mathematical}. For a head $h$, the attention weight $\alpha_{i,j}^{h}$ acts as a modulation coefficient controlling the intensity of information flow from the source position $j$ to the target position $i$. Given this, we focus on measuring the importance of attention weights to identify semantic and structural pathways. Unlike traditional causal mediation analysis~\cite{wang2022interpretability,hanna2023does}, CAA introduces counterfactual contrast at the group level, eliminating the reliance on strict token correspondence of original and counterfactual inputs. CAA includes three steps: estimating the sample-wise importance; deriving the group-wise importance via span aggregation; introducing the contrastive importance score to identify pathways.

\paragraph{Sample-wise Importance.}
Given the input $x=(t, q)\in\mathcal{D}$ and a task metric $m$,  we measure the importance of $\alpha_{i,j}^{h}$ via the indirect effect~\cite{Pearl:2001:DIE:2074022.2074073,vig2020investigating}:
\begin{equation}
I(m, \alpha_{i,j}^{h}, x) = m(x) - m(x \mid \operatorname{do}(\alpha_{i,j}^{h}=0)),
\label{eq:patch}
\end{equation}
where $\operatorname{do}(\alpha_{i,j}^{h}=0)$ zeros out $\alpha_{i,j}^{h}$. Eq.~\eqref{eq:patch} describes the induced change for $m$ under zero ablation. Based on it, we define two new metrics for semantic checking $m_\text{sem}$ and structural matching $m_\text{str}$,  respectively:
\begin{equation}
m_\text{sem}(x) = \max_{w \in \mathcal{R}} \log P(w|x) - \log P(w_\text{tool}|x),
\label{eq:sem_metric}
\end{equation}
\begin{equation}
m_\text{str}(x) = \log P(w_\text{tool}|x) - \max_{w \in \mathcal{R}} \log P(w|x).
\label{eq:str_metric}
\end{equation}
Here, $\mathcal{R}=\{w_{r_1}, \dots, w_{r_k}\} \subset \mathcal{V}$ is a set of refusal tokens that the model typically starts with when refusing invocations (See Appendix~\ref{app:special_tokens} for a discussion on refusal token selection.). Due to the large number of attention weights, we further use attribution patching~\cite{neelattribution} to efficiently approximate Eq.~\eqref{eq:patch} via a first-order Taylor expansion:
\begin{equation}
I(m, \alpha_{i,j}^{h}, x) \approx \alpha_{i,j}^{h} \cdot \frac{\partial m(x)}{\partial \alpha_{i,j}^{h}}.
\end{equation}
This enables parallel estimation for all $\alpha$ using only one forward and one backward pass.

\paragraph{Group-wise Importance.}
Since sequence lengths vary across samples, the importance of attention weights is not directly comparable.
To address this, we adopt a span-level aggregation approach~\cite{haklay-etal-2025-position} by partitioning the input text into ordered spans $\mathcal{S}=\langle s_i \rangle_{i=1}^{N}$, where the tool definition and the query are treated as separate spans to ensure consistency across samples. Details of the span partitioning are provided in Appendix~\ref{app:span}.
Consequently, the group-wise importance from a source span $s_{j}$ to a target span $s_{i}$, measured over $\mathcal{D}$, is defined as:
\begin{equation}
\hat{I}^h_{s_{i}, s_{j}}(m, \mathcal{D}) = \frac{1}{|\mathcal{D}|}\sum_{x\in \mathcal{D}} \sum_{l \in s_j} \sum_{k \in s_i} I(m, \alpha_{k,l}^{h}, x).
\label{eq:span_agg}
\end{equation}

\paragraph{Contrastive Importance Score.}
Based on Eq.~\eqref{eq:patch}, we can identify attention weights that contribute significantly to the metric $m$. However, this manner captures the aggregated contributions of all features originating from $x$, making it difficult to disentangle the effects of specific factors.
In our setting, we are specifically interested in the effects induced by semantic irrelevance and structural alignment.
To isolate these factors, we construct counterfactual samples and introduce \textbf{Contrastive Importance Score} (CIS).
Let $\mathcal{D}^* = \{x^* \mid x \in \mathcal{D}\}$ be the set of counterfactual samples corresponding to $\mathcal{D}$. We define CIS as:
\begin{align}
\mathrm{CIS}^h_{s_i,s_j}(m, \mathcal{D}, \mathcal{D}^*) 
&= 
Z^h_{s_i, s_j}(m, \mathcal{D}) - \nonumber \\
& \quad Z^h_{s_i, s_j}(m, \mathcal{D}^*),
\label{eq:cis}
\end{align}
\begin{equation}
Z^h_{s_i, s_j}(m, \mathcal{D}') = \frac{\max\left(\hat{I}^h_{s_i, s_j}(m, \mathcal{D}'),0\right)} {\max_{{h}', s_p, s_q} \hat{I}^{h'}_{{s_p}, s_q}(m, \mathcal{D}')}.
\label{eq:normal}
\end{equation} 
Here, Eq.~(\ref{eq:normal}) is a normalized form of group-wise importance, where $\mathcal{D}'$ denotes either $\mathcal{D}$ or $\mathcal{D}^*$. In Eq.~\eqref{eq:cis}, by differencing two normalized terms, we remove the effects of features shared in $x$.
\par
CAA constructs two types of counterfactuals. As shown in Table~\ref{tab:intervention_examples}, we construct structural counterfactuals $\mathcal{D}^{*, \mathrm{str}}$ by parameter substitution, and semantic counterfactuals $\mathcal{D}^{*, \mathrm{sem}}$ by pairing a query with its target tool:
\begin{equation}
\mathrm{CIS}^{h,\text{sem}}_{s_{i},s_{j}} 
\coloneqq 
\mathrm{CIS}^h_{s_{i},s_{j}}(m_\text{sem}, \mathcal{D}_\text{nocall}, \mathcal{D}^{*, \text{sem}}_\text{nocall}),
\label{eq:sem_cis}
\end{equation}
\begin{equation}
\mathrm{CIS}^{h,\text{str}}_{s_{i},s_{j}} 
\coloneqq 
\mathrm{CIS}^h_{s_{i},s_{j}}(m_\text{str}, \mathcal{D}_\text{call}, \mathcal{D}^{*, \text{str}}_\text{call}).
\label{eq:str_cis}
\end{equation}
Here, $\mathcal{D}_\text{nocall}$ and $\mathcal{D}_\text{call}$ consist of cases where the model originally refuses or performs tool invocations, and decisions on their respective counterfactuals, $\mathcal{D}_\text{nocall}^{*, \mathrm{sem}}$ and $\mathcal{D}_\text{call}^{*, \mathrm{str}}$, are swapped\footnote{For example, $\mathcal{D}_\text{nocall}$ is constructed by selecting samples where the model refuses the original (irrelevant) tool and invokes the tool after replacing it with the target tool.}. We compute Eq.~\eqref{eq:sem_cis} and Eq.~\eqref{eq:str_cis}, and identify the top-$k$ attention weights between spans as semantic and structural pathways, respectively.

\subsection{Experiment \& Analysis}
\label{sec:mech_result}
We split SABEval into training/validation/test sets (40\%, 20\%, 40\%), with disjoint tool templates.
We compute CIS by sampling 500 instances per group ($\mathcal{D}_\mathrm{nocall}$ and $\mathcal{D}_\mathrm{call}$) from the training split of the $\mathcal{D}_1$ subset.
All analyses are performed on the held-out test set.
Unless otherwise specified, we select the top-$k$ attention weights where $k$ corresponds to 2\% of the number of attention heads.

\begin{figure}[t]
    \centering
    \includegraphics[width=0.494\linewidth]{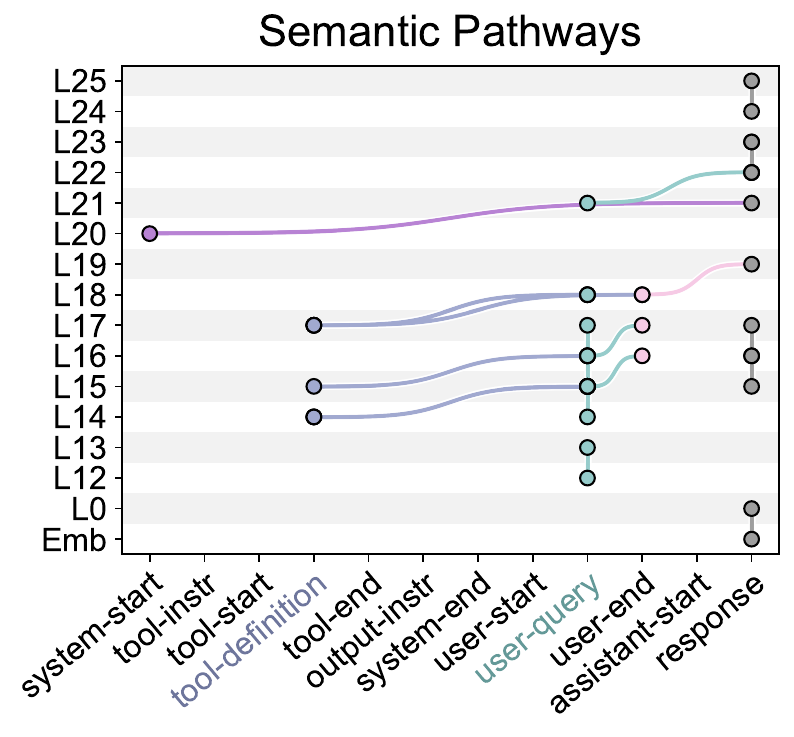}
    \includegraphics[width=0.494\linewidth]{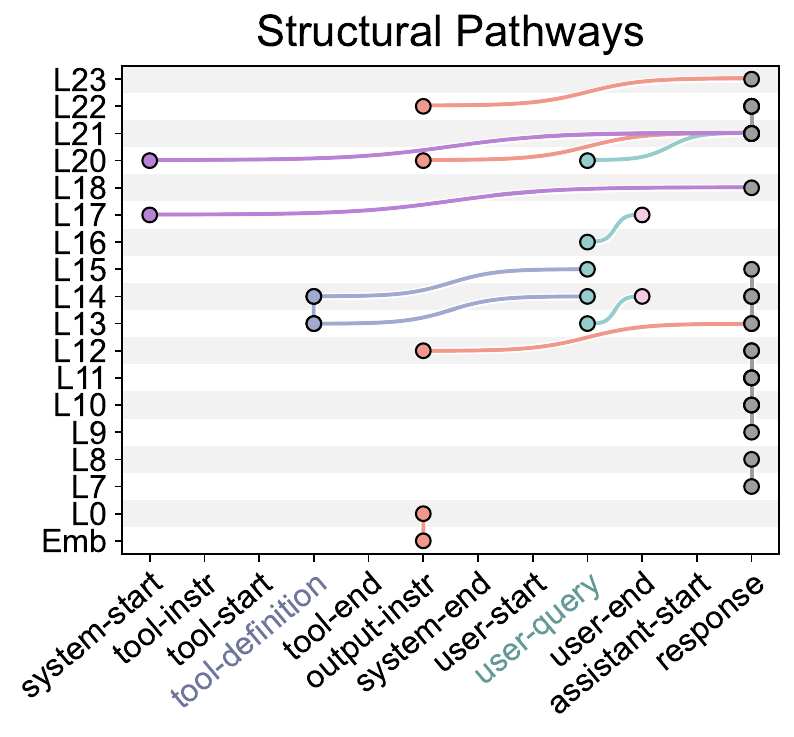}
    \caption{Semantic (left) and structural (right) pathways in Qwen3-8B from top-$k$ attention weights ($k=23$, corresponding to 2\% of the attention heads).}
    \label{fig:mech_pathways_qwen3_8b}
\end{figure}

\paragraph{Characteristics of Pathways.}
Fig.~\ref{fig:mech_pathways_qwen3_8b} illustrates the semantic and structural pathways identified in Qwen3-8B.
The x-axis represents the ordered input spans (e.g., tool definition, user query), and the y-axis represents the model layer. Edges denote top-$k$ attention weights aggregated across heads between source and target spans at different layers.
While both pathways involve similar layers, we observe clear differences in their information flow patterns.
Specifically, semantic pathways are denser at the query positions, indicating that these positions play a key role during semantic checking.
In contrast, during structural matching, the model exhibits strong attention at the final span toward the output instruction, which specifies the tool invocation format. This pattern suggests that excessive focus on output formatting biases Qwen3 toward structural alignment.
More discussion and results for other models are provided in Appendix~\ref{app:pathway_ch}.

\begin{figure}[t]
  \centering
  \includegraphics[width=\columnwidth]{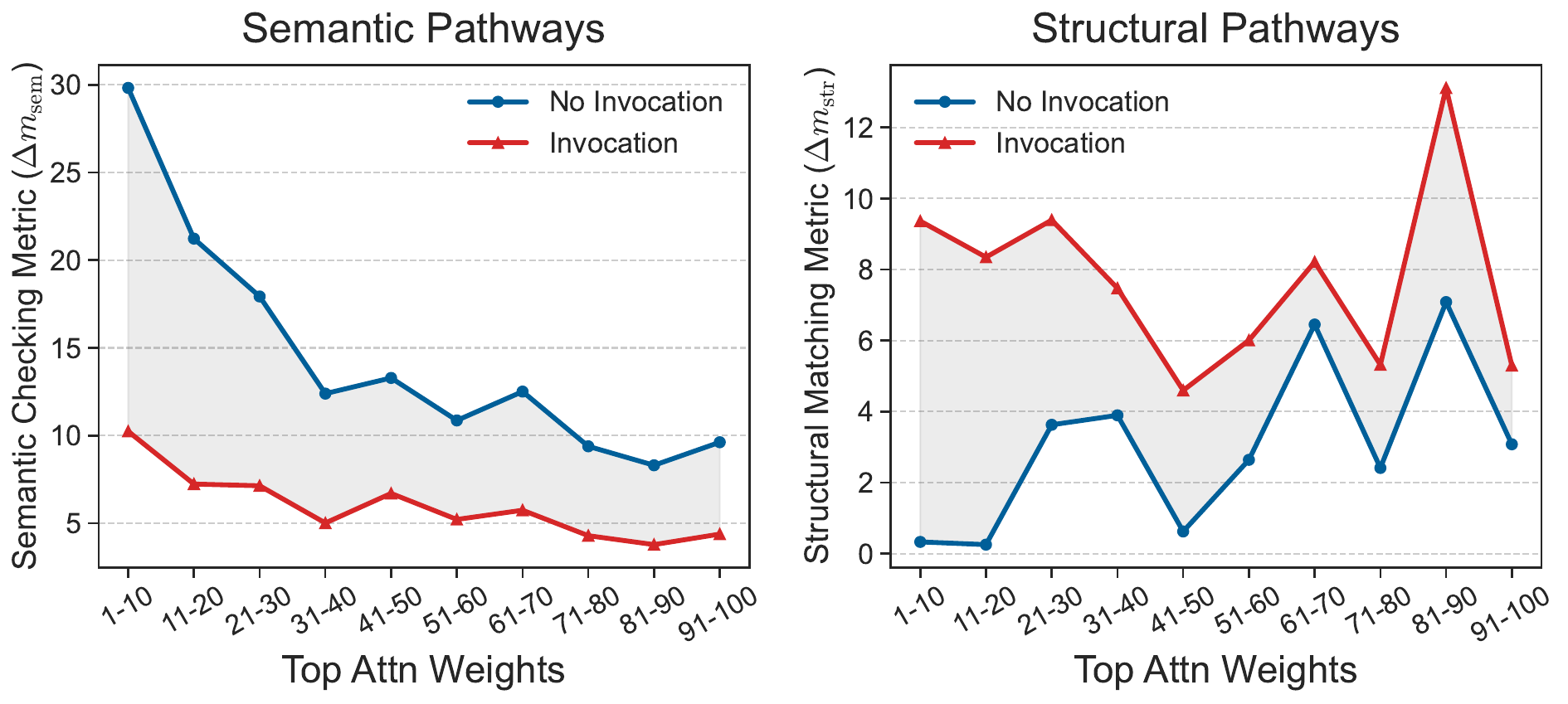}
  \caption{Comparison of pathway strengths between invocation and non-invocation cases for Qwen3-8B.}
  \label{fig:mech_compare}
\end{figure}

\paragraph{Relative Pathway Strength Drives Model Decisions.}
We find that tool invocation decisions are governed by the relative strength of the two pathways.
To quantify this, we partition the top-100 attention weights into 10 groups and iteratively patch each group via Eq.~\eqref{eq:patch} to assess their impacts on $m_{\text{sem}}$ and $m_{\text{str}}$, using 500 invocation and 500 non-invocation cases from $\mathcal{D}_\text{call}$ and $\mathcal{D}_\text{nocall}$ of $\mathcal{D}_1$, respectively.
As shown in Fig.~\ref{fig:mech_compare}, pathway strengths vary significantly: erroneous invocation cases exhibit weaker semantic pathways and stronger structural pathways, opposite to non-invocation cases.
We further show in Appendix~\ref{app:compare_degree} that as structural alignment degree increases, structural pathways strengthen while semantic pathways weaken, consistent with the higher TIR noted in \S~\ref{subsec:phenomenon}.
These results indicate that the relative dominance of the semantic and structural pathways determines LLMs' tool invocation decisions.

\begin{table*}[h]
\centering
\footnotesize
\renewcommand{\arraystretch}{0.85}
\setlength{\aboverulesep}{1pt}
\setlength{\belowrulesep}{2pt}
\begin{tabular}{l l M{1.1cm} M{1.1cm} M{1.1cm} M{1.1cm} M{1.1cm} M{1.1cm} M{1.1cm}}
\toprule
\multirow{2}{*}{\textbf{Model}} & \multirow{2}{*}{\textbf{Method}} & \multicolumn{5}{c}{\textbf{SABEval $\bm{\downarrow}$}} & \multicolumn{2}{c}{\textbf{ACEBench $\bm{\uparrow}$}} \\
\cmidrule(lr){3-7} \cmidrule(lr){8-9}
 & & \textbf{$\bm{\mathcal{D}_0}$} & \textbf{$\bm{\mathcal{D}_1}$} & \textbf{$\bm{\mathcal{D}_2}$} & \textbf{$\bm{\mathcal{D}_3}$} & \textbf{$\bm{\mathcal{D}_4}$} & \textbf{Single} & \textbf{Parallel} \\
\midrule
\multirow{3}{*}{Qwen3-4B} & Base & 37.61 & 57.85 &  73.41 & 85.56 & 89.76 & 65 & 60 \\
 & Prompt & 27.12 & 42.20 & 56.83 & 69.71 & 75.95 & \textbf{66} & 57 \\
 & \gc CAA (Ours) & \gc \textbf{8.29} & \gc \textbf{14.29} & \gc \textbf{20.29} & \gc \textbf{27.76} & \gc \textbf{36.00} & \gc 63 & \gc \textbf{61} \\ 
\midrule

\multirow{3}{*}{Qwen3-8B} & Base & 30.39 & 51.61 & 66.29 & 75.56 & 82.44 & 71 & 62 \\
 & Prompt & 19.90 & 34.29 & 48.73 & 57.66 & 66.20 & \textbf{72} & \textbf{66} \\
 & \gc CAA (Ours) & \gc \textbf{4.29} & \gc \textbf{6.49} & \gc \textbf{7.27} & \gc \textbf{8.98} & \gc \textbf{11.12} & \gc 67 & \gc \textbf{66} \\ 
\midrule

\multirow{3}{*}{Qwen3-14B} & Base & 41.37 & 59.46 & 74.73 & 83.90 & 91.02 & 68 & 62 \\
 & Prompt & 24.29 & 33.41 & 43.56 & 56.78 & 65.66 & \textbf{70} & 62 \\
 & \gc CAA (Ours) & \gc \textbf{5.56} & \gc \textbf{7.27} & \gc \textbf{9.31} & \gc \textbf{15.51} & \gc \textbf{21.22} & \gc 68 & \gc \textbf{63} \\ 
\midrule

\multirow{3}{*}{ToolACE-2.5} & Base & 37.95 & 51.12 & 63.02 & 67.41 & 68.00 & \textbf{90} & \textbf{81} \\
 & Prompt & 39.12 & 53.66 & 62.63 & 67.71 & 67.71 & 87 & \textbf{81} \\
 & \gc CAA (Ours) & \gc \textbf{3.41} & \gc \textbf{5.27} & \gc \textbf{5.76} & \gc \textbf{6.88} & \gc \textbf{11.46} & \gc 89 & \gc 79 \\ 
\midrule

\multirow{3}{*}{Watt-Tool} & Base & 8.68 & 15.46 & 23.31 & 31.51 & 39.51 & \textbf{84} & 76 \\
 & Prompt & 8.59 & 15.37 & 22.34 & 29.61 & 38.05 & 83 & \textbf{77} \\
 & \gc CAA (Ours) & \gc \textbf{2.54} & \gc \textbf{2.98} & \gc \textbf{5.46} & \gc \textbf{8.10} & \gc \textbf{12.68} & \gc 83 & \gc 75 \\ 

\bottomrule
\end{tabular}
\caption{Evaluation results on all subsets of SABEval and ACEBench. We report the tool invocation rate ($\downarrow$) on SABEval and accuracy ($\uparrow$) on ACEBench. Base denotes the performance of the original model.}
\label{tab:main_result}
\end{table*}

\begin{figure}[t]
  \centering
  \includegraphics[width=\columnwidth]{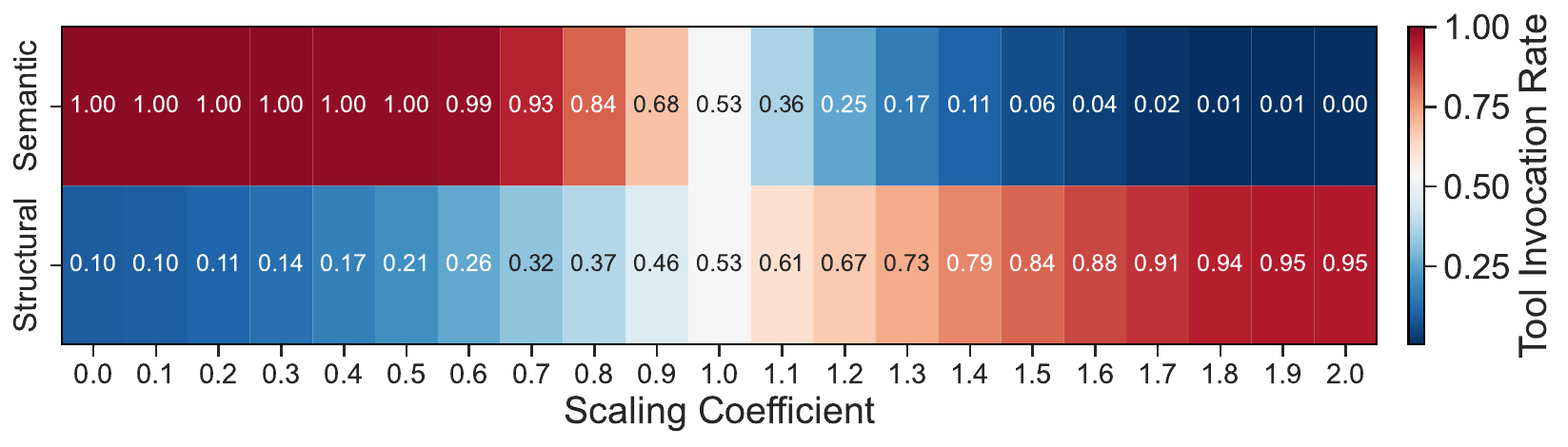}
  \caption{TIR under different scaling coefficients for Qwen3-8B ($k=23$).}
  \label{fig:mech_scale_single}
\end{figure}

\paragraph{Mitigating the Bias via Attention Scaling.}
If these pathways indeed govern LLMs' decision process, manipulating their strength should causally influence models' behavior.
To validate their causal roles, we intervene on them by scaling the attention weights of head $h$ between span $s_i$ and $s_j$:
\begin{equation}
\alpha^{h}_{k,l} \leftarrow \rho \cdot \alpha^{h}_{k,l}, \quad \forall k \in s_i, l \in s_j, \end{equation}
where $\rho$ is a scaling coefficient.
We first examine how TIR varies with different $\rho$ on 500 test cases.
As shown in Fig.~\ref{fig:mech_scale_single}, the two pathways exert causal effects on the tool invocation rate (TIR) in opposite directions, which is consistent with their hypothesized roles and demonstrates the expected causal effect.
Results for other models and different top-$k$ thresholds are reported in Appendix~\ref{app:pathway_intervention},
which show similar trends.

Motivated by these findings, we propose a targeted intervention strategy to mitigate structural alignment bias.
Specifically, we amplify semantic pathways ($\rho_{\text{sem}} > 1$) while suppressing structural pathways ($\rho_{\text{str}} < 1$).
The coefficients $\rho_{\text{sem}}$ and $\rho_{\text{str}}$ are selected on the validation set (See Appendix~\ref{app:coefficient} for details.).
We evaluate efficacy on SABEval and general tool-use performance on ACEBench (single and parallel settings\footnote{The model must correctly select one or simultaneously multiple tools, and then fulfill the desired parameters. We report the accuracy metric.})~\cite{chen2025acebench}, and compare our method against the prompt baseline\footnote{Prompting the model to refuse the invocation when no suitable tool is available. See Appendix~\ref{app:prompt_baseline} for details. We also compare with one-shot prompting and supervised fine-tuning in Appendix~\ref{app:additional_baselines}.}.
\par
The results are shown in Table~\ref{tab:main_result}, highlighting three key observations:
(1) Our method substantially reduces structural alignment bias, achieving a 45.55\% average reduction in tool invocation rate, whereas the prompt baseline shows more limited improvements (11.19\% average reduction).
(2) With disjoint tool templates, rebalancing the pathways discovered from tools in the training set remains effective in mitigating bias on samples generated from unseen tools. This indicates that the pathways are not tool-specific but generalize across tools.
(3) Our method has negligible impact on general tool-use capabilities. Results show that our method introduces only minimal accuracy changes on ACEBench, which are comparable to those introduced by prompting.

\section{Related Work}
\paragraph{Irrelevant Tools.}
Prior studies have explored scenarios involving irrelevant tools. \citet{xu2024reducing} frame tool irrelevance as a form of task insolvability and introduce a dedicated ``ChangeTools action'', while \citet{huang2023metatool, ross-etal-2025-when2call, patil2025the} require LLMs to explicitly acknowledge when no tool is applicable. Other works further expect models to clarify irrelevance or address it through user interaction~\cite{chen2025acebench, liu-etal-2025-abstain, wang-etal-2025-learning}. However, these works ignore the factor of structural alignment, thereby obscuring the underlying bias.

\paragraph{Irrelevant Context.}
Although irrelevant tools appear as part of an LLM's context, they differ fundamentally from irrelevant context studied in prior works~\cite{cuconasu2024power, wu2024easily, niu-etal-2025-llama, cheng-etal-2025-stochastic}. First, irrelevant tools change the model's expected behavior (tool invocation to refusal), whereas irrelevant context does not.
Moreover, despite prior work identifying the competition between irrelevant context and internal knowledge~\cite{cheng-etal-2025-stochastic}, the competing pathways found in our work arise entirely from irrelevant tools themselves, reflecting two distinct query–tool matching mechanisms.

\paragraph{Mechanistic Interpretability.}
Mechanistic interpretability aims to reverse-engineer the computations of LLMs to uncover the causal mechanisms behind their behaviors~\cite{Olah2022, wang2022interpretability}.
Recent studies indicate that contextual information is often utilized by specific attention heads~\cite{wu2024retrieval, jin-etal-2024-cutting,crosbie-shutova-2025-induction}, employing causal mediation analysis that relies on counterfactual interventions on model components~\cite{vig2020investigating,meng2022locating,syed2024attribution,haklay-etal-2025-position}.
By extending this framework, we gain an internal mechanistic understanding of the causes of structural alignment bias and provide the mitigation method.

\section{Conclusion}
In this work, we identify structural alignment bias, which can lead LLMs to erroneously invoke irrelevant tools, potentially resulting in serious and irreversible consequences.
To study this problem, we construct a new dataset and conduct a comprehensive empirical analysis. Our results indicate that structural alignment bias is prevalent, can dominate tool-refusal behavior, and substantially increases erroneous invocations as structural alignment strengthens. Moreover, we introduce CAA to link this behavioral pattern to competing internal signals. Building on these insights, we further design a rebalancing strategy that mitigates this bias without sacrificing general tool-use capabilities.

\section*{Limitations}
To rigorously isolate structural alignment from semantic relevance, we focus on the irrelevance setting in tool invocations and therefore do not extend our analysis to multi-turn agentic workflows. Nevertheless, understanding this fundamental bias in isolation provides a critical prerequisite for addressing it in more complex settings. Moreover, although we systematically analyze structural alignment bias, we do not distinguish whether it originates from general pre-training or is specifically introduced during tool-use fine-tuning. We leave the precise pinpointing of its source to future work.

\section*{Acknowledgments}
This work is supported by the National Natural Science Foundation of China (No.62402491, No.U2336202), the China Postdoctoral Science Foundation (No.2025M771524), and the Shanghai Pujiang Program (24PJA004).

\bibliography{custom}

\clearpage
\newpage
\appendix

\section{Dataset Construction}
\label{app:dataset}

By leveraging the polymorphism principle of Object-Oriented Programming~\cite{meyer1997object}, we construct groups of sibling tools that share an identical interface (i.e., parameters) but possess mutually exclusive functionalities (i.e., they belong to different derived classes), enabling us to create samples that are structurally aligned yet semantically irrelevant.
In what follows, we detail the construction procedure in three steps:

\paragraph{Hierarchical Tool Construction.}
We select seed tools from two mainstream benchmarks: BFCL~\cite{patil2025the} and ACEBench~\cite{chen2025acebench}.
First, we sanitize these tools and convert them into a unified format.
To minimize parsing noise and focus on decision logic, we filter out seed tools that contain nested (i.e., object type) parameters and mandate all parameters as required. Subsequently, we manually annotate a \textit{base class} for each seed tool. These base classes are derived typically from the entities targeted by the seed tools (e.g., mapping ``Basketball Player'' to ``Athlete''). Then, we annotate \textit{tool templates} by replacing specific entity references in the seed tools with a generic placeholder, denoted as ``\texttt{<class>}''.

Based on tool functionality, we then annotate \textit{derived classes} for each base class. We enforce two main constraints during this process:

(1) The seed tool interface must be directly applicable to every derived class.

(2) The derived classes must be semantically mutually exclusive, such that a tool instantiated for one derived class cannot resolve queries intended for another.

Finally, to instantiate specific derived tools, we simply replace the ``\texttt{<class>}'' placeholder with the target derived class name via string substitution. This process yields tools that share identical parameters but possess mutually exclusive functionalities.
We provide an example in Fig.~\ref{fig:dataset_tool_example}.

\paragraph{Query Generation.}
For each instantiated derived tool, we leverage GPT-4o~\cite{gpt4o} to generate diverse user queries. To ensure the quality and validity of the data, we mainly enforce three constraints during the generation process:

(1) The query must explicitly incorporate information corresponding to every parameter defined in the tool.

(2) The query must explicitly mention the name of the derived class associated with the tool.

(3) The query must be unambiguous and fully self-contained, requiring no further clarification to be understood.

The specific prompt template used for this generation is illustrated in Fig.~\ref{fig:dataset_query_generation}.

\paragraph{Sibling Pairing.}
Sibling tools are those derived from the same tool template (i.e., sharing the same base class) but instantiated with distinct derived classes.
Finally, we construct samples $(t, q)$ by pairing a query, originally generated from a specific tool, with one of that tool's sibling tools.
Through this pairing strategy, the pair $(t, q)$ exhibits structural alignment because the tool accommodates every attribute present in the query, yet it remains semantically irrelevant.

\paragraph{Data Statistics.}
Since the seed tools are sourced from benchmarks with broad coverage, SABEval naturally inherits the domain diversity. Specifically, our dataset spans a wide spectrum of topics, including the seven domains explicitly identified in ACEBench: finance, society, health, entertainment, technology, environment, and culture~\cite{chen2025acebench}.

\begin{figure}[t]
    \centering
    \includegraphics[width=0.6775\linewidth]{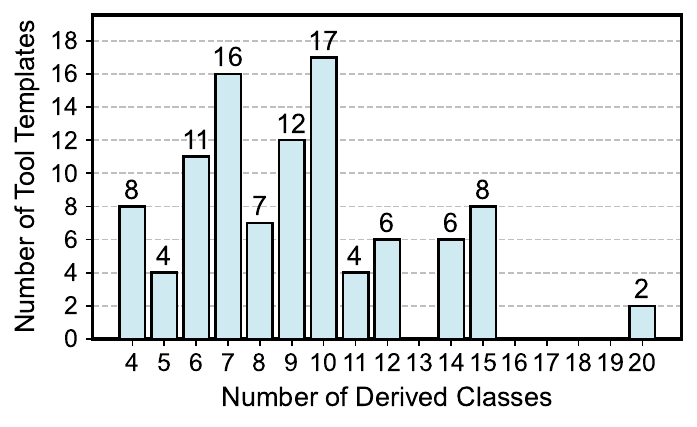}
    \includegraphics[width=0.31\linewidth]{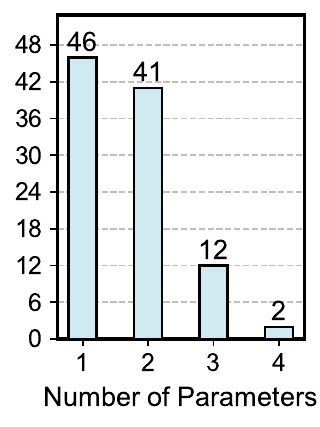}
    \caption{Distribution of derived class counts (left) and parameter counts (right) across tool templates.}
    \label{fig:data_stats}
\end{figure}

In total, SABEval ($\mathcal{D}_0$) comprises 101 tool templates, each defined with 1 to 4 parameters.
Across the base classes corresponding to these templates, we annotated a total of 729 distinct derived classes.
On average, each tool template is instantiated into 9 derived tools, with at least 5 queries generated for each derived tool. We sample 10 distinct sibling pairs per tool template to construct the final samples.
Fig.~\ref{fig:data_stats} shows the distributions of derived class counts and parameter counts across tool templates.

\paragraph{Dataset Extensions.}
As described in \S~\ref{sec:dataset}, we construct four dataset extensions, $\mathcal{D}_1, \dots, \mathcal{D}_4$, by augmenting the tool templates from $\mathcal{D}_0$ with $k \in \{1, 2, 3, 4\}$ additional parameters, respectively. We employ GPT-4o to synthesize these supplementary parameters, followed by a manual verification process to ensure their validity. The specific prompt used for this process is provided in Fig.~\ref{fig:dataset_extension}.

\section{System Prompts}
\label{sec:apd_models}

The default system prompts for all models are shown in Figs.~\ref{fig:qwen_prompt},~\ref{fig:toolace_prompt}.

\begin{figure*}[t]
  \centering
  \includegraphics[width=0.32\linewidth]{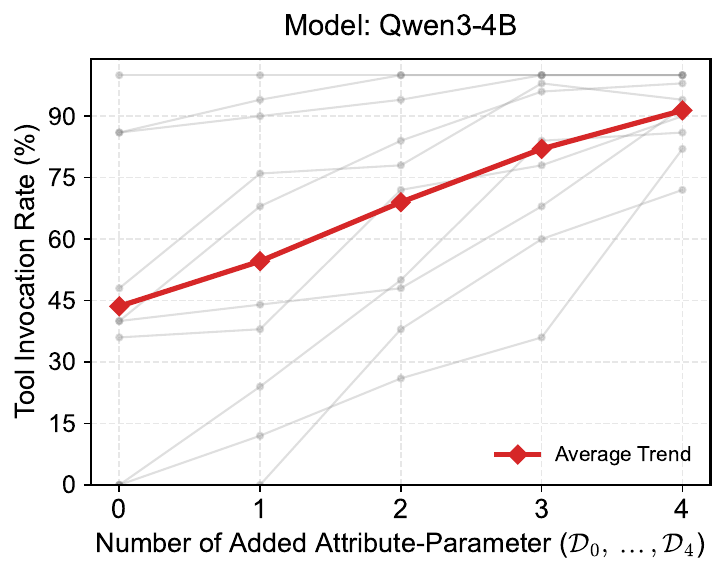}
  \includegraphics[width=0.32\linewidth]{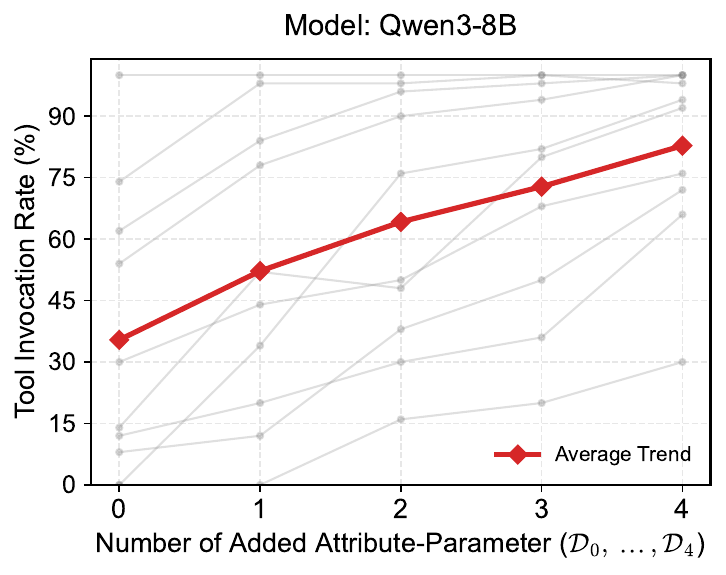}
  \includegraphics[width=0.32\linewidth]{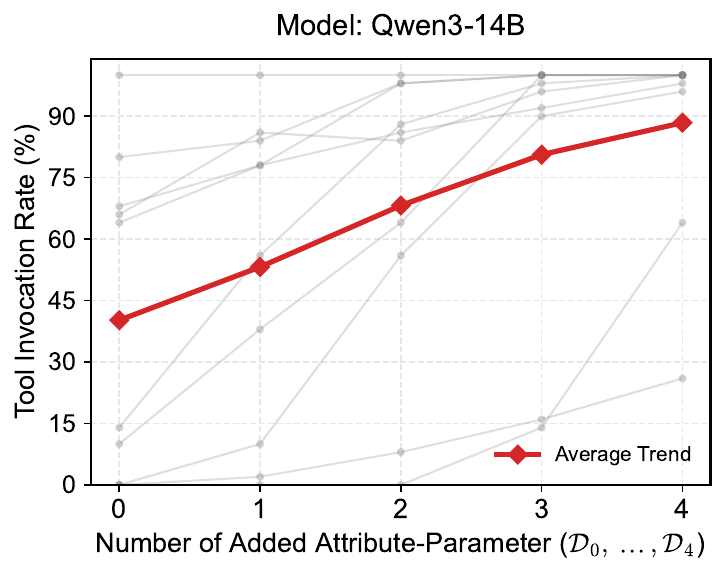}

  \includegraphics[width=0.32\linewidth]{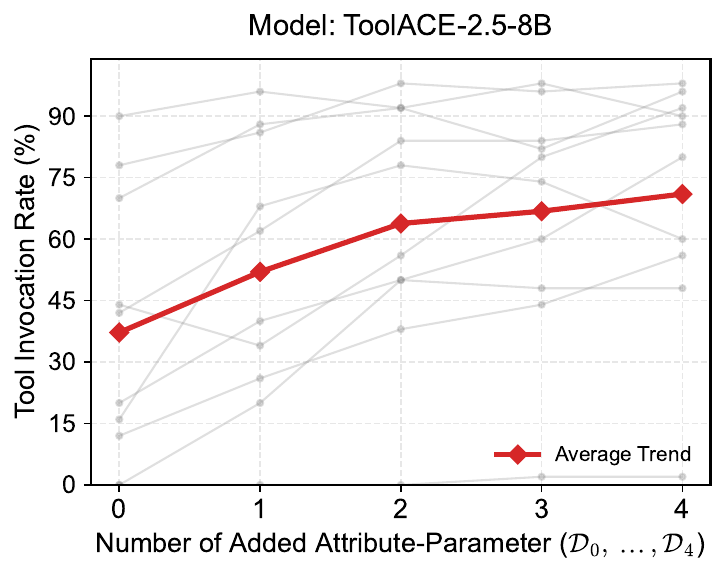}
  \includegraphics[width=0.32\linewidth]{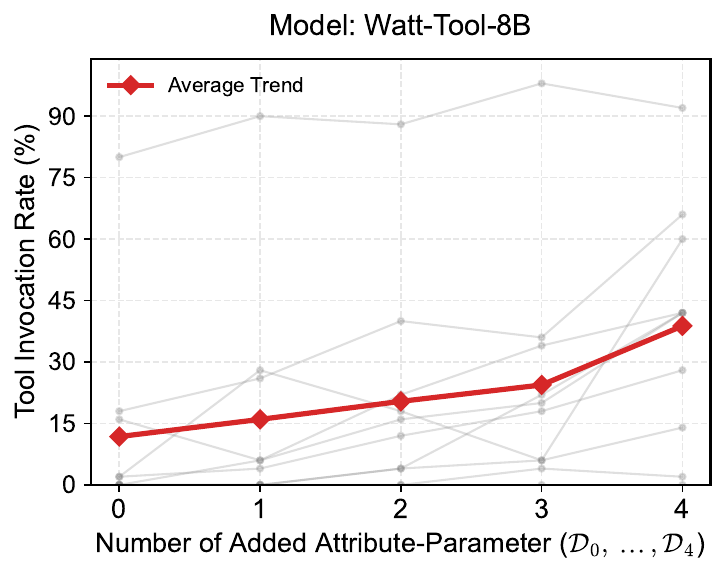}
  \caption{Tool invocation rate (\%) across 10 randomly sampled tool templates. The \textcolor{red}{red} line denotes the overall TIR aggregated over these 10 groups, while the \textcolor{gray!90}{gray} lines represent individual tool template groups.}
  \label{fig:group_tir}
\end{figure*}

\section{TIR Analysis Across Tool Templates}
\label{app:group_tt}
We randomly sample 10 tool templates from SABEval and compute the average TIR on the samples generated by each template, as well as the overall TIR aggregated across all 10 templates. As shown in Fig.~\ref{fig:group_tir}, the TIR exhibits a highly consistent increasing trend with respect to the degree of structural alignment across different tool templates.

\section{Probability Decrease in Structural Counterfactuals}
\label{app:prob_shift}
We investigate the probability shifts of $w_{\text{tool}}$ in structural counterfactuals. 
A decrease in $P(w_{\text{tool}})$ directly reflects a reduced tendency for tool invocation. 
Specifically, we count samples for which the probability drops by more than 5\% relative to the original probability, i.e., 
$P(w_{\text{tool}} \mid q, t^*) < 0.95 \cdot P(w_{\text{tool}} \mid q, t)$.
As shown in Table~\ref{tab:prob_cf}, over 70\% of counterfactual samples exhibit such decreases for the Qwen3 series.
This proportion further increases to 90.65\% and 96.45\% for ToolACE-2.5-8B and Watt-Tool-8B, respectively.
These results further substantiate the strong causal effect of structural alignment on tool invocations.

\begin{table}[h]
  \centering
  \small
  \setlength{\extrarowheight}{1.5pt}
  \begin{tabular}{p{2.2cm}M{1.2cm}M{1.2cm}M{1.2cm}}
    \toprule
    \textbf{Model} & \textbf{Add} & \textbf{Remove} & \textbf{Substitute} \\
    \midrule
    Qwen3-4B        & 38.81 & 68.57 & \textbf{70.30} \\
    Qwen3-8B        & 42.65 & 72.67 & \textbf{76.71} \\
    Qwen3-14B       & 42.43 & 66.48 & \textbf{73.76} \\
    ToolACE-2.5-8B  & 81.92 & 80.55 & \textbf{90.65} \\
    Watt-Tool-8B    & 86.63 & 91.05 & \textbf{96.45} \\
    \bottomrule
  \end{tabular}
  \caption{Percentage of samples in which the probability of $w_{\text{tool}}$ drops by more than 5\% relative to its original value under structural counterfactuals.}
  \label{tab:prob_cf}
\end{table}

\begin{table}[h]
\centering
\small
\resizebox{\columnwidth}{!}{%
\begin{tabular}{l c c c c c}
\toprule
\multirow{2}{*}{\textbf{Models}} & \multirow{2}{*}{\textbf{Original}} & \multicolumn{4}{c}{\textbf{Instantiate Param Count}} \\

\cmidrule(lr){3-6} 

 & & 1 & 2 & 3 & 4 \\
\midrule

\cellcolor{gray!10}$\operatorname{sim}(t,q)$ & \cellcolor{gray!10}52.26 & \cellcolor{gray!10}54.72 & \cellcolor{gray!10}57.48 & \cellcolor{gray!10}60.57 & \cellcolor{gray!10}63.09 \\
\midrule
Qwen3-4B             & 89.76 & 87.39 & 81.56  & 73.50  & 65.11 \\
Qwen3-8B             & 83.05 & 77.64 & 71.60  & 64.30  & 57.07 \\
Qwen3-14B              & 90.44 & 88.10 & 82.14  & 77.96  & 70.99 \\
ToolACE-2.5            & 71.47 & 69.66 & 70.00 & 66.71  & 63.09 \\
Watt-Tool              & 39.98 & 38.93 & 34.42  & 31.39  & 24.87 \\
\bottomrule
\end{tabular}}
\caption{TIR for cases flattened with different numbers of parameters (from $\mathcal{D}_4$ of SABEval). The similarity score $\operatorname{sim}(t, q)$ is also reported.}
\label{tab:rewrite}
\end{table}

\section{Discussion about Textual Similarity}
\label{app:text_sim}
Compared to the structural counterfactual samples obtained in \S~\ref{sec:causal}, the original structurally aligned samples exhibit more shared concepts and fewer semantic conflicts. Consequently, they naturally yield higher semantic textual similarity\footnote{Commonly measured by cosine similarity of dense vector representations, e.g., Sentence-BERT~\cite{reimers-gurevych-2019-sentence}.}. This observation is intuitive, as textual similarity can be viewed as a downstream consequence of structural alignment.

\begin{table*}[h]
    \centering
    \small
    \begin{tabular}{ll}
        \toprule
        \multicolumn{2}{l}{\cellcolor{gray!10}\textbf{User Query:} I need \textbf{hiking shoes} recommendations for \textcolor{entityRed}{males} dealing with \textcolor{entityBlue}{overpronation}.} \\
        \multicolumn{2}{l}{\cellcolor{gray!10}\textbf{Tool Name:} \textbf{sandals}\_prep} \\
        \midrule
        \multirow{2}{*}{\textbf{Original}} & \textbf{Tool Description:} Provides a specific list of \textbf{sandals} recommendations for common foot conditions. \\
         & \textbf{Tool Parameters:} \textcolor{entityBlue}{\texttt{foot\_condition}}, \textcolor{entityRed}{\texttt{gender}} \\
        \midrule
        \multirow{2}{*}{\textbf{Flattened}} & \textbf{Tool Description:} Provides a specific list of \textcolor{entityRed}{male} \textbf{sandals} recommendations for common foot conditions. \\
         & \textbf{Tool Parameters:} \textcolor{entityBlue}{\texttt{foot\_condition}} \\
        \bottomrule
    \end{tabular}
    \caption{An illustration of the structurally flattened case.}
    \label{fig:flatten}
\end{table*}

However, we argue that structural alignment implies more than just elevated textual similarity; specifically, textual similarity alone is insufficient to explain the observed bias. We demonstrate this through a \textit{structural flattening} experiment, where we move target parameters from the tool definition to the tool description by instantiating them with query attributes (see Table.~\ref{fig:flatten}).
This operation increases the naive textual similarity between the tool and the query, as the tool description now contains exact mentions of the query attributes.
We then measure the cosine similarity between the tool and the query embeddings\footnote{We use EmbeddingGemma-300M~\cite{vera2025embeddinggemma}.} and evaluate TIR.

As shown in Table~\ref{tab:rewrite}, instantiating more parameters improves textual similarity; however, TIR declines as LLMs need to predict fewer parameters, suggesting that merely increasing textual similarity does not consistently drive up TIR. This indicates that the structural alignment bias is systematically induced by the structured nature of the data rather than naive textual similarity.

\section{Special Tokens}
\label{app:special_tokens}

\paragraph{Tool-call Tokens.}
When invoking a tool, LLMs are instructed to start their responses with a special token $w_{\text{tool}}$, which enables us to analyze the tool-invocation behavior.
The concrete $w_{\text{tool}}$ for each model and their output formats are shown in Table~\ref{tab:tool_call_tokens}.

We also inspect the full generated texts on $\mathcal{D}_0$ of SABEval and find no cases where LLMs emit this token without invoking a provided tool (i.e., hallucinating a non-existent tool).

\paragraph{Refusal Tokens.}
Consistent with observations of \citet{arditi2024refusal}, we find that model generations typically start with specific phrases when rejecting a tool. Therefore, for each model, we define a set of refusal tokens $\mathcal{R}=\{w_{r_1}, \dots, w_{r_k}\} \subset \mathcal{V}$, and then define the semantic checking metric $m_\text{sem}$ and structural matching metric $m_\text{str}$ in Eq.~\eqref{eq:sem_metric} and Eq.~\eqref{eq:str_metric}.
The concrete refusal tokens we choose for each model are shown in Table~\ref{tab:refusal_tokens}. These tokens cover over 98\% of refusal cases.

We use the log-probability difference between the tool-call token and refusal tokens, rather than the former alone, for two reasons. First, when a model is highly confident, absolute probabilities saturate near 0 or 1. In this Softmax region, gradients approach zero, causing attribution patching (Eq.~\ref{eq:patch}) to severely underestimate the causal impact of attention heads and potentially leading to numerical instability. Second, the difference metric isolates the competition between invocation and refusal by canceling out the Softmax denominator, eliminating confounding effects from other tokens. For example, if the model assigns high probability to a malformed tool invocation, the probability of the tool-call token would decrease due to the unit-sum constraint, conflating formatting errors with genuine refusal.

\begin{table*}[t]
\centering
\small
\begin{tabular}{m{2.2cm}M{2.5cm}M{3cm}m{5.7cm}}
    \toprule
    \textbf{Model}  &  \textbf{Tool-call token} &  \textbf{Tool-call token string} & \textbf{Tool-call format} \\
    \midrule
    Qwen3 family           & 151657 & ``\texttt{\textcolor{blue}{<tool\_call>}}'' & 
    \texttt{\textcolor{blue}{<tool\_call>}\newline\{"name": <func-name>, "arguments": <args-json-object>\}\newline</tool\_call>}     \\
    \midrule
    ToolACE-2.5-8B  & 58     & ``\texttt{\textcolor{blue}{[}}''   & \texttt{\textcolor{blue}{[}func\_1(param\_a=value\_a, param\_b=value\_b...),func\_2(params)]} \\
    \midrule
    Watt-Tool-8B    & 58     & ``\texttt{\textcolor{blue}{[}}''   & \texttt{\textcolor{blue}{[}func\_1(param\_a=value\_a, param\_b=value\_b...),func\_2(params)]} \\
    \bottomrule
\end{tabular}
\caption{Tool-call tokens $w_{\text{tool}}$, their string representations, and concrete tool-call formats}
\label{tab:tool_call_tokens}
\end{table*}

\begin{table*}[t]
\centering
\small
\begin{tabular}{m{2.2cm}M{2.5cm}M{3cm}m{5.7cm}}
    \toprule
    \textbf{Model}  &  \textbf{Refusal tokens} &  \textbf{Refusal tokens string} & \textbf{Example refusal} \\
    \midrule
    Qwen3 family           & \{785, 40, 4064, 2132\} & \{"The", "I", "None", "It"\} & 
    ``\textcolor{blue}{It} seems there might be a misunderstanding. The provided function is for comparing VR headsets, not drones. If you'd like to compare VR headsets......''     \\
    \midrule
    ToolACE-2.5-8B  & \{791, 40\} & \{"The", "I"\} & ``\textcolor{blue}{I}'m sorry, but I can only provide the current price for a game on the PlayStation platform. I do not have access to GOG prices.''\\
    \midrule
    Watt-Tool-8B    & \{791, 40\} & \{"The", "I"\} & ``\textcolor{blue}{The} given question lacks the parameters required by the function. The available function is for basketball, not lacrosse.'' \\
    \bottomrule
\end{tabular}
\caption{Refusal tokens for each model, corresponding string representations, and an example refusal. These tokens cover over 98\% of refusal cases.}
\label{tab:refusal_tokens}
\end{table*}

\section{Span Partitioning}
\label{app:span}
We partition the input text into contiguous and non-overlapping spans to measure group-wise importance. The specific partitions for different models are shown in Tables~\ref{tab:span_qwen3}, \ref{tab:span_toolace}.

\section{Semantic and Structural Pathways}
\label{app:pathway_ch}
\begin{figure*}[h]
    \centering
    \includegraphics[width=0.45\linewidth]{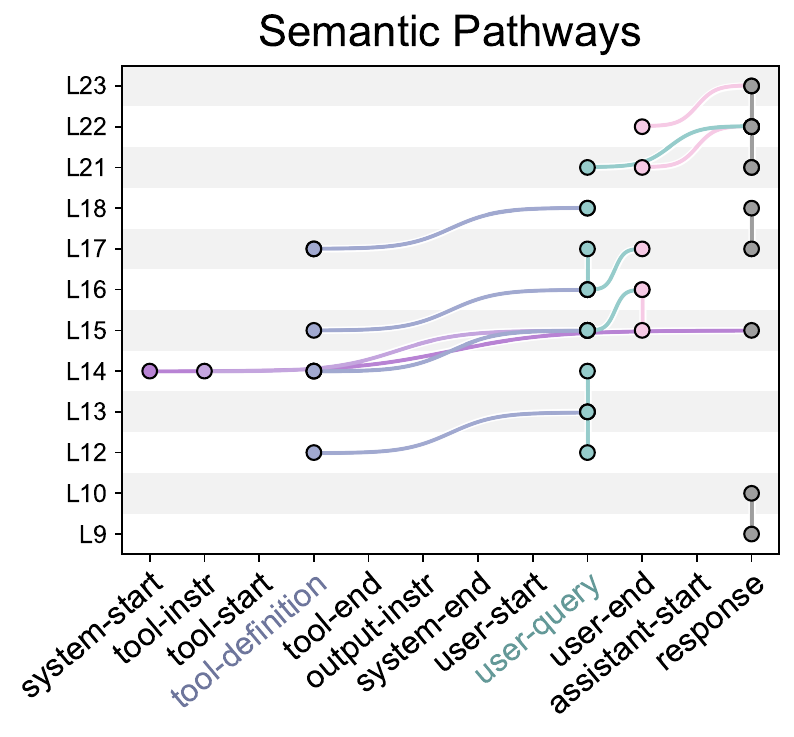}
    \includegraphics[width=0.45\linewidth]{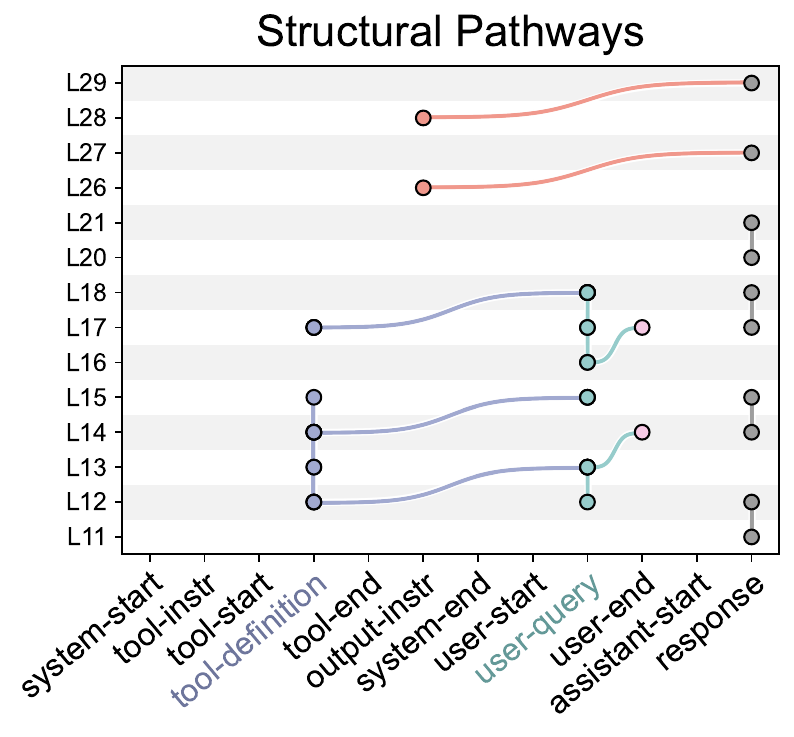}
    \caption{Pathways in Qwen3-4B.}
    \label{fig:mech_pathways_qwen3_4b}
\end{figure*}

\begin{figure*}[h]
    \centering
    \includegraphics[width=0.45\linewidth]{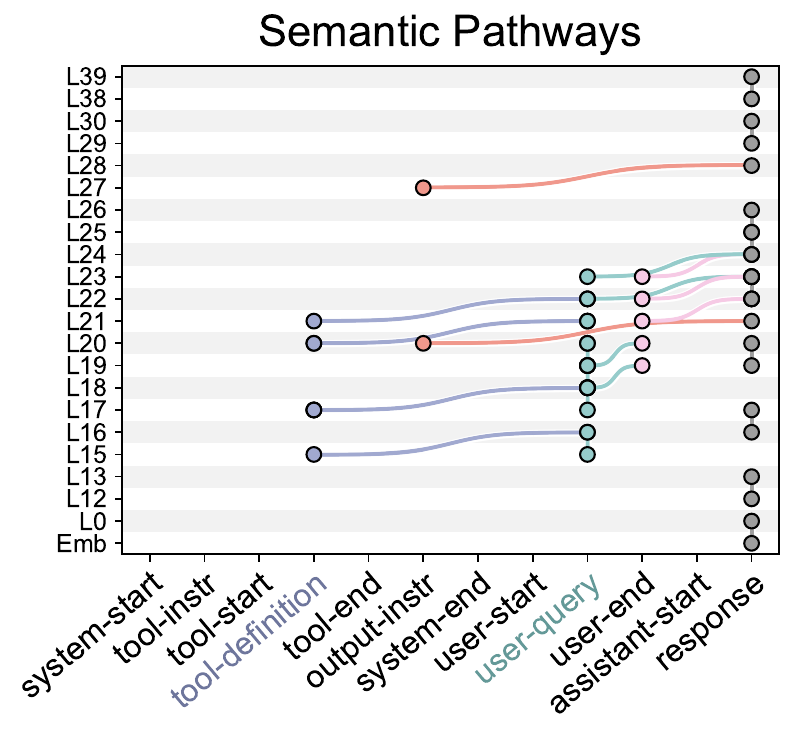}
    \includegraphics[width=0.45\linewidth]{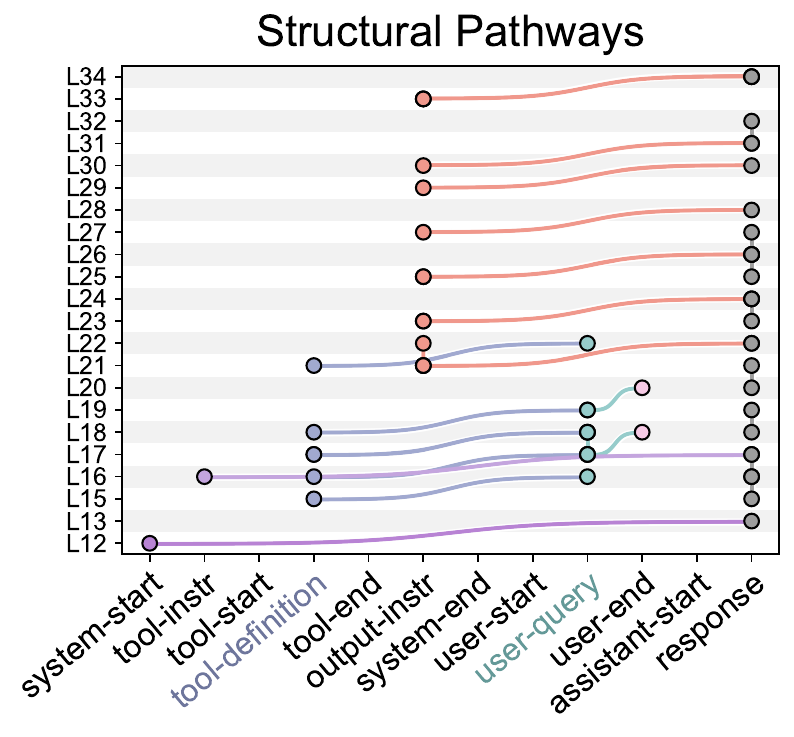}
    \caption{Pathways in Qwen3-14B.}
    \label{fig:mech_pathways_qwen3_14b}
\end{figure*}

\begin{figure*}[h]
    \centering
    \includegraphics[width=0.45\linewidth]{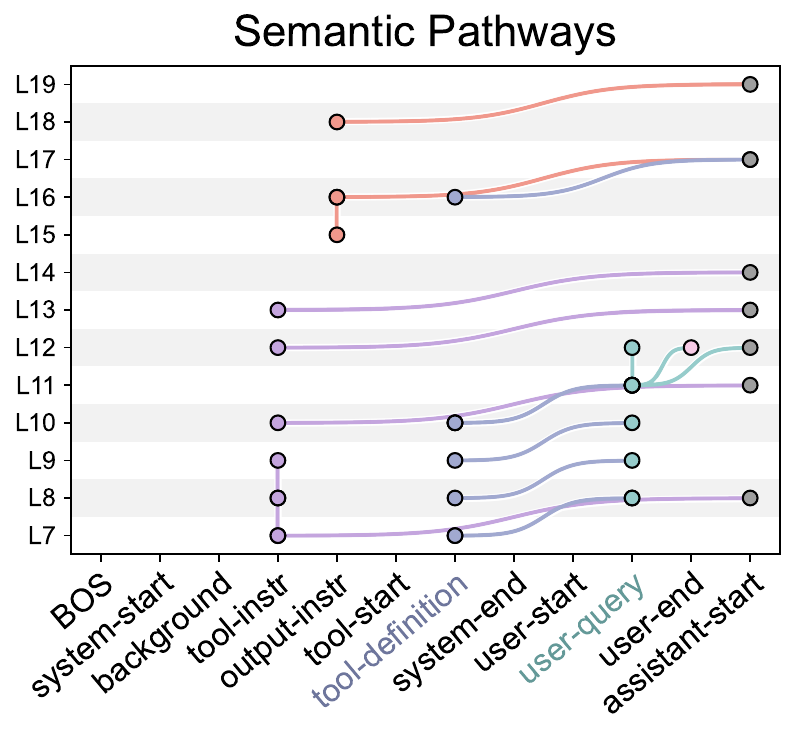}
    \includegraphics[width=0.45\linewidth]{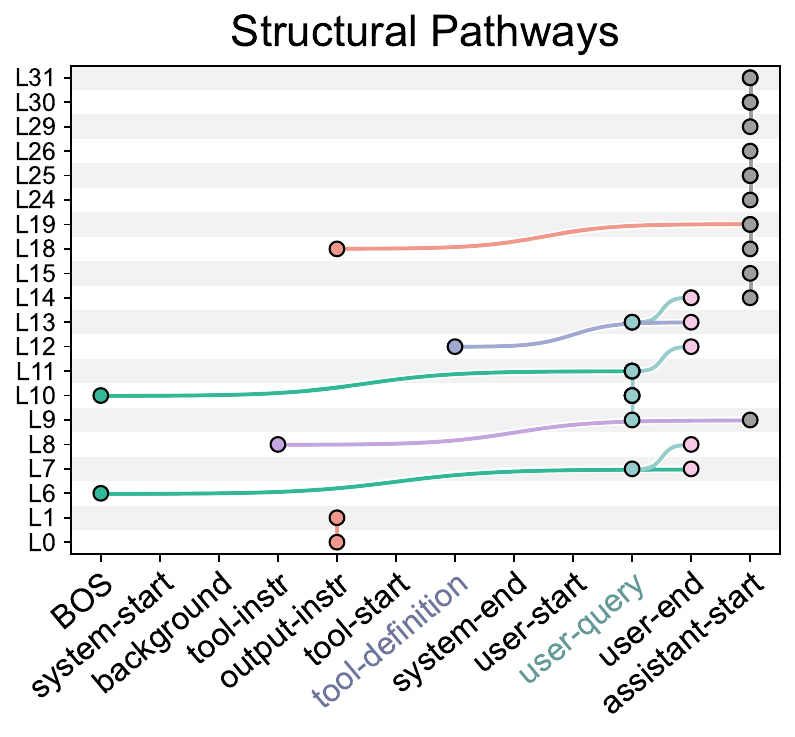}
    \caption{Pathways in ToolACE-2.5-8B.}
    \label{fig:mech_pathways_toolace}
\end{figure*}

\begin{figure*}[h]
    \centering
    \includegraphics[width=0.45\linewidth]{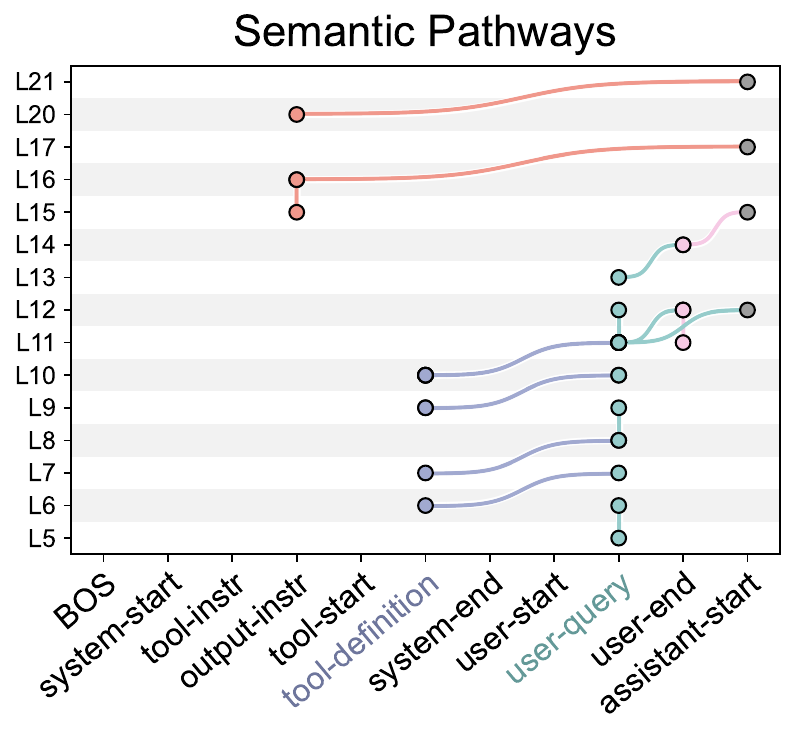}
    \includegraphics[width=0.45\linewidth]{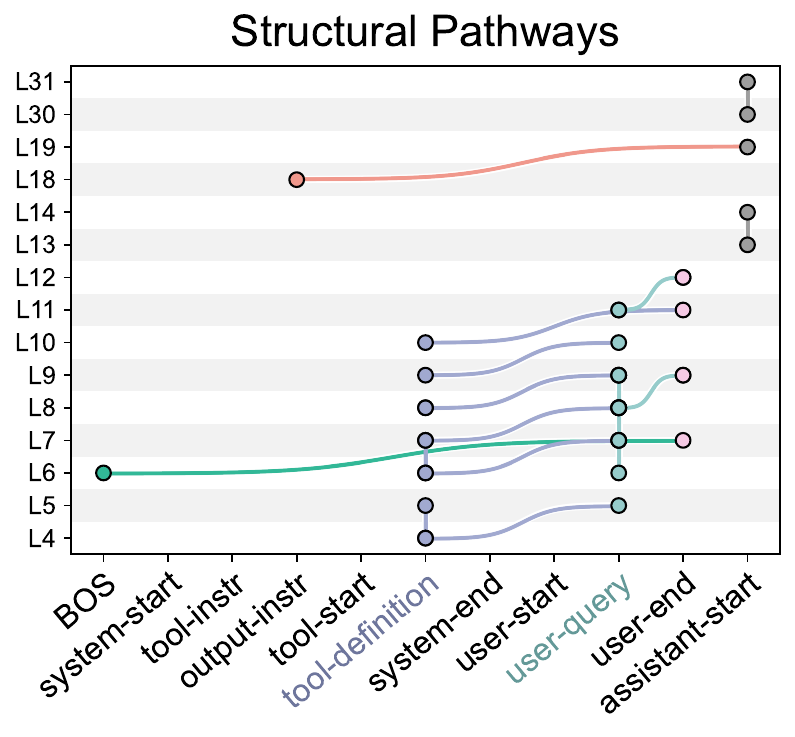}
    \caption{Pathways in Watt-Tool-8b.}
    \label{fig:mech_pathways_watt}
\end{figure*}

We visualize the identified pathways for other models in Figs.~\ref{fig:mech_pathways_qwen3_4b},~\ref{fig:mech_pathways_qwen3_14b},~\ref{fig:mech_pathways_toolace},~\ref{fig:mech_pathways_watt}.

Across all models, tool information primarily flows to the query positions and subsequently to the last span (either directly or via the end position of user message). The Qwen3-4B and 14B models exhibit characteristics consistent with the 8B version.
However, ToolACE-2.5-8B and Watt-Tool-8B diverge in their pathway distributions. Specifically, ToolACE-2.5-8B shows denser semantic pathways at the tool positions, whereas Watt-Tool-8B exhibits more focused structural pathways on these positions, and neither model concentrates its structural pathways on the tool output format. Moreover, for ToolACE-2.5-8B, the semantic pathways heavily attend to the tool instruction, which explicitly outlines tool rejection, suggesting a dependency on this prompt during semantic checking.

\section{Pathway Strengths Across Degrees of Structural Alignment}
\label{app:compare_degree}
\begin{figure*}[t]
  \centering
  \includegraphics[width=0.9\linewidth]{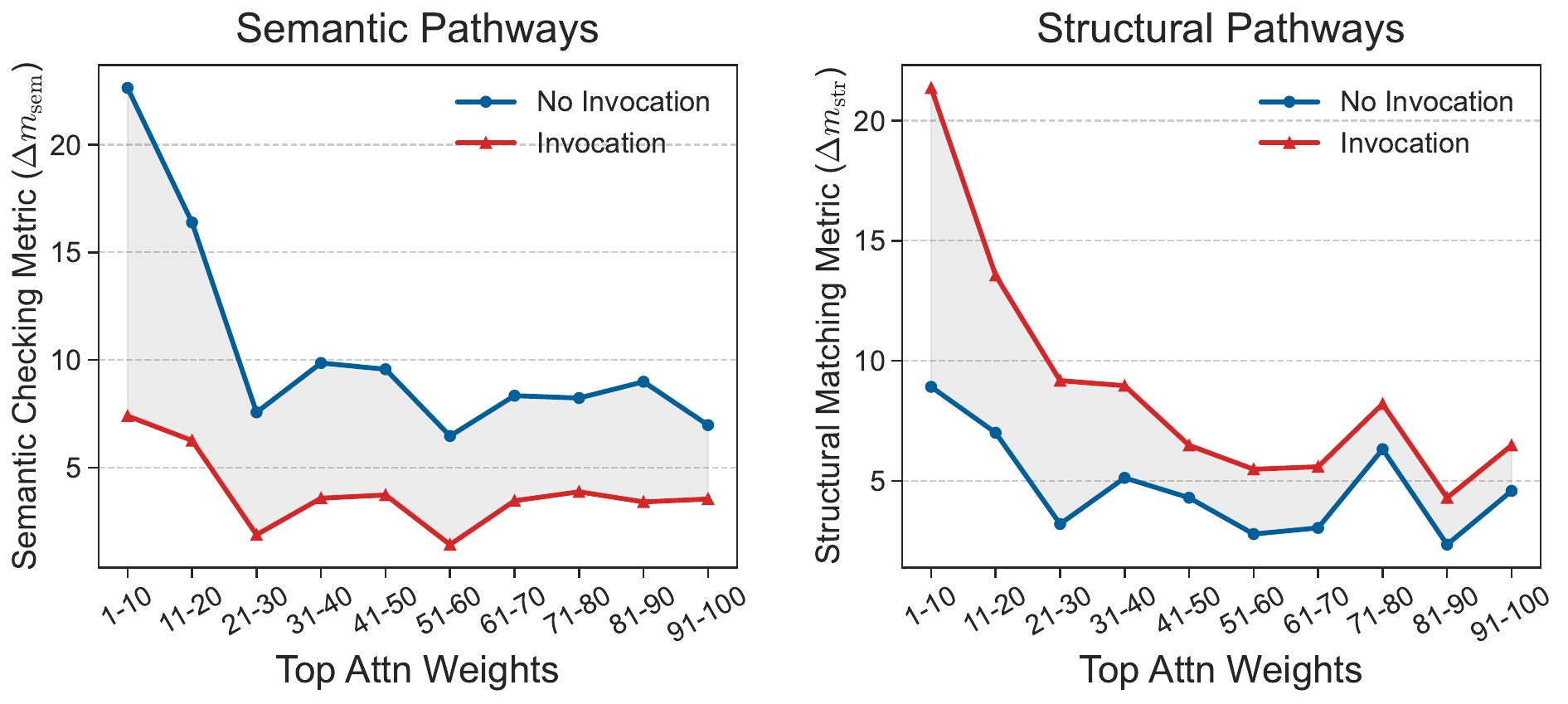}
  \caption{Comparison of pathway strengths between invocation and non-invocation cases for Qwen3-4B.}
  \label{fig:app_compare_qwen3_4b}
\end{figure*}

\begin{figure*}[t]
  \centering
  \includegraphics[width=0.9\linewidth]{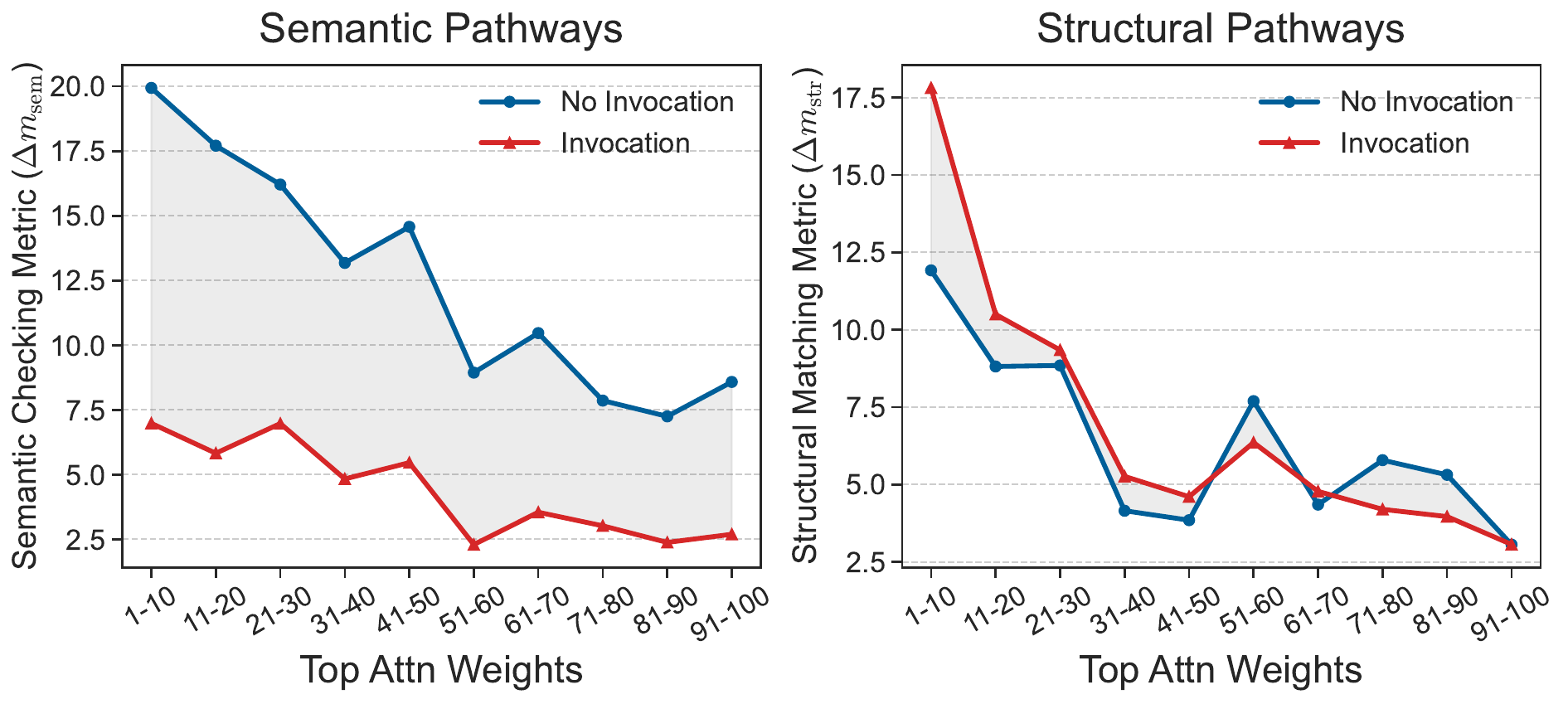}
  \caption{Comparison of pathway strengths between invocation and non-invocation cases for Qwen3-14B.}
  \label{fig:app_compare_qwen3_14b}
\end{figure*}

\begin{figure*}[t]
  \centering
  \includegraphics[width=0.9\linewidth]{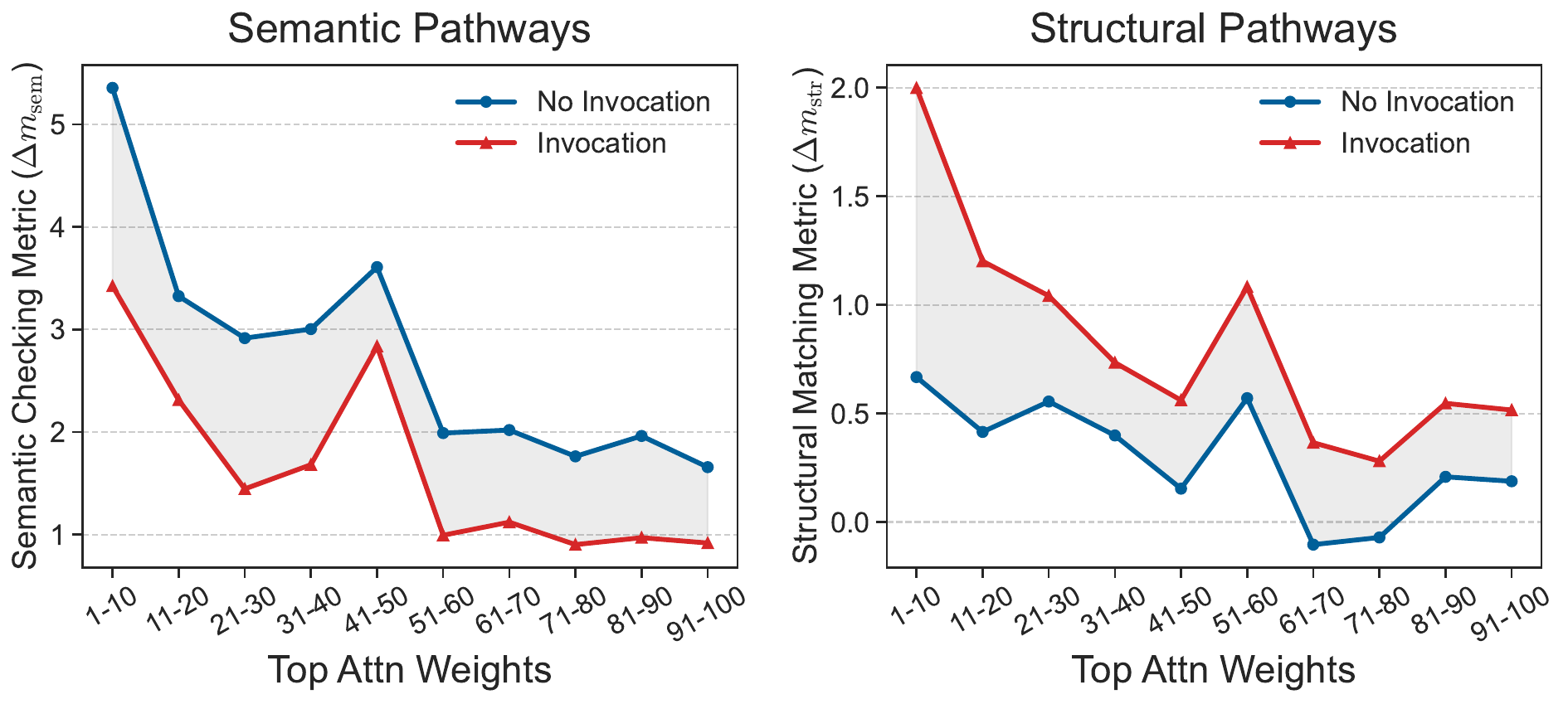}
  \caption{Comparison of pathway strengths between invocation and non-invocation cases for ToolACE-2.5-8B.}
  \label{fig:app_compare_toolace}
\end{figure*}

\begin{figure*}[t]
  \centering
  \includegraphics[width=0.9\linewidth]{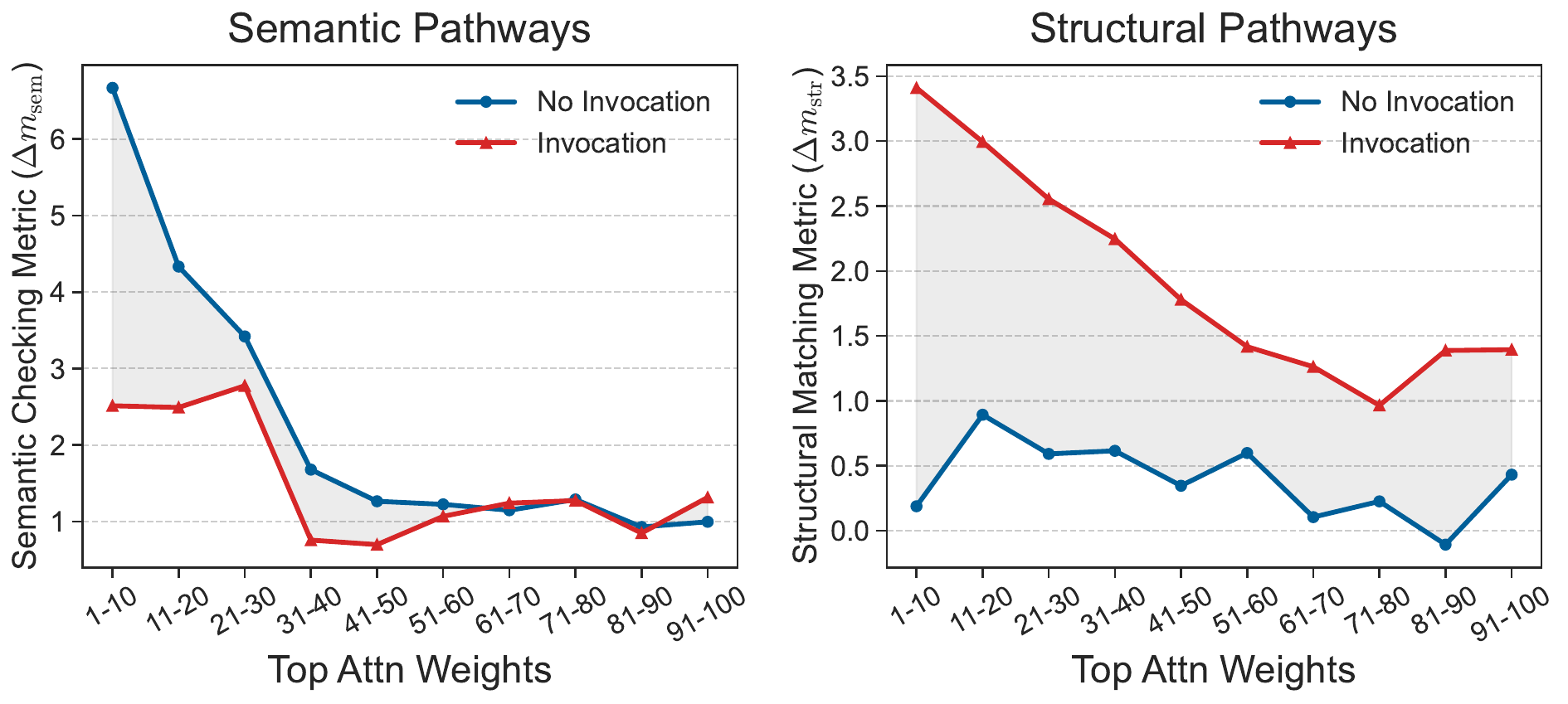}
  \caption{Comparison of pathway strengths between invocation and non-invocation cases for Watt-Tool-8B.}
  \label{fig:app_compare_watt}
\end{figure*}

We present the comparison of pathway strengths between invocation and non-invocation cases for other models in Figs.~\ref{fig:app_compare_qwen3_4b},~\ref{fig:app_compare_qwen3_14b},~\ref{fig:app_compare_toolace},~\ref{fig:app_compare_watt}.

\begin{figure*}[h]
    \centering
    \includegraphics[width=0.32\linewidth]{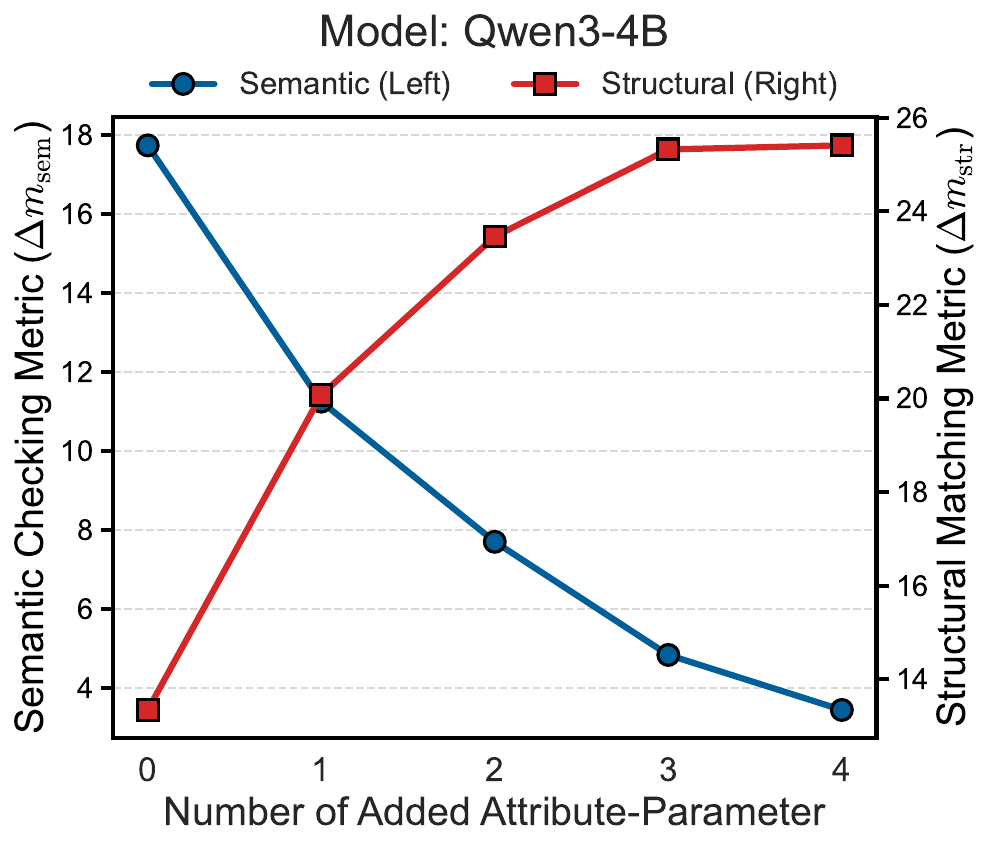}
    \includegraphics[width=0.32\linewidth]{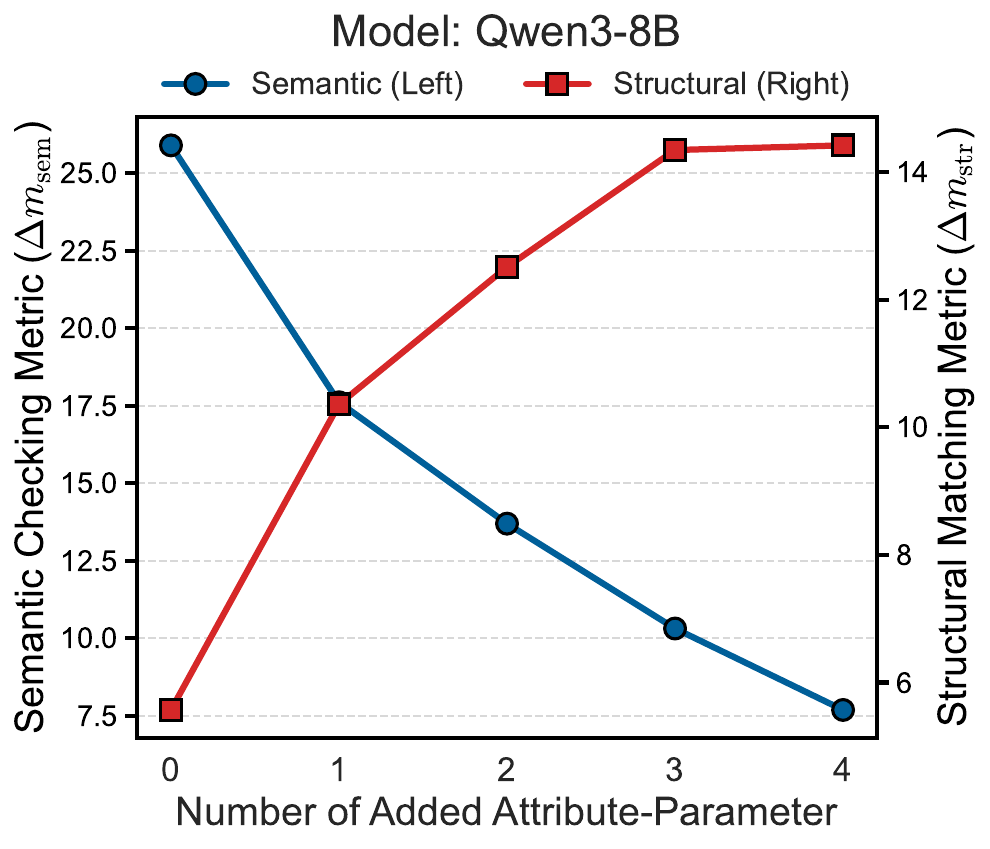}
    \includegraphics[width=0.32\linewidth]{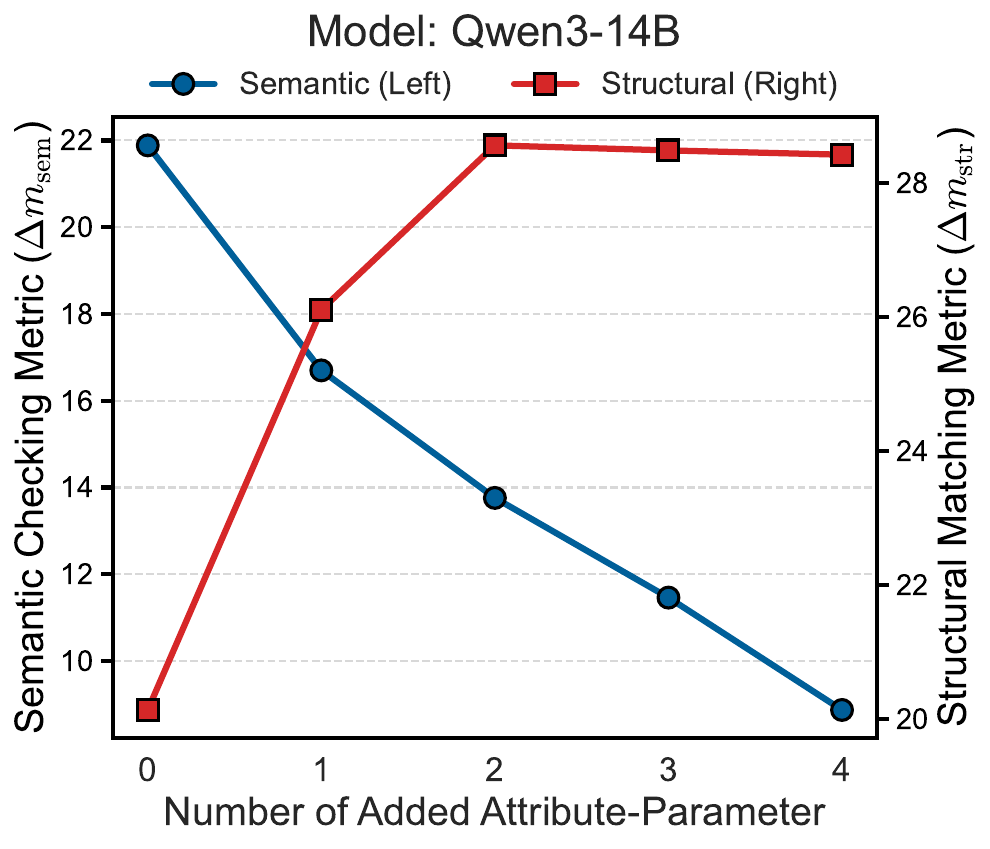}
    
    \includegraphics[width=0.32\linewidth]{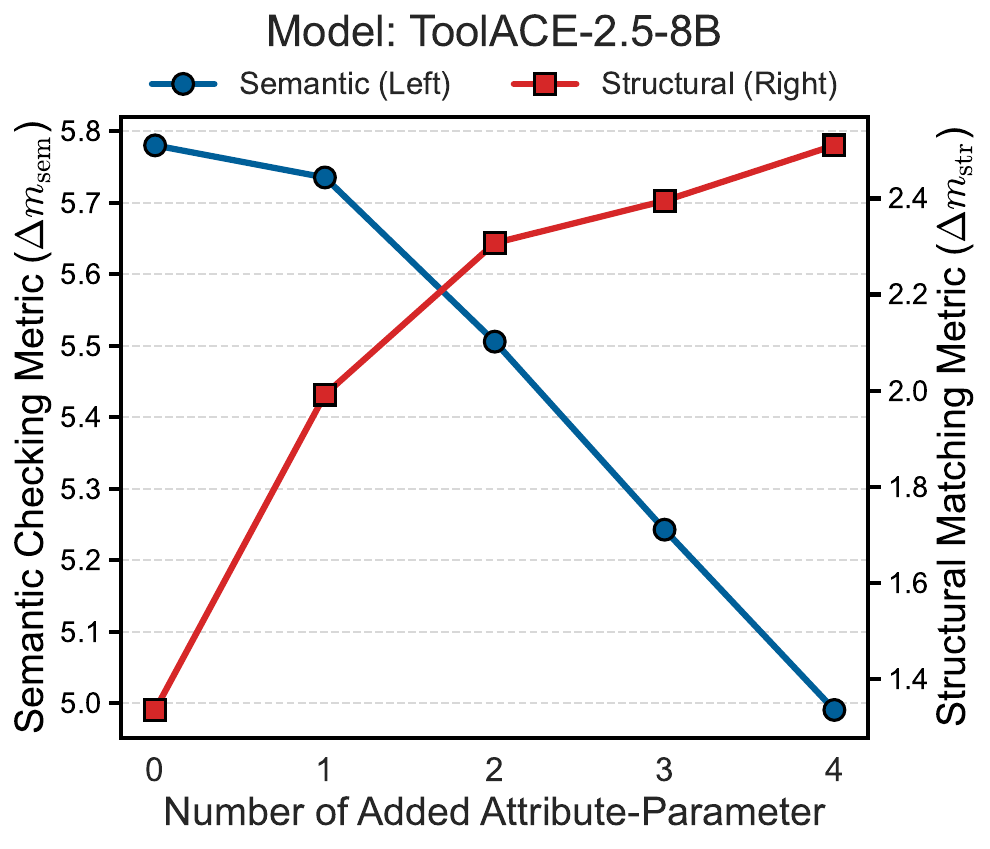}
    \includegraphics[width=0.32\linewidth]{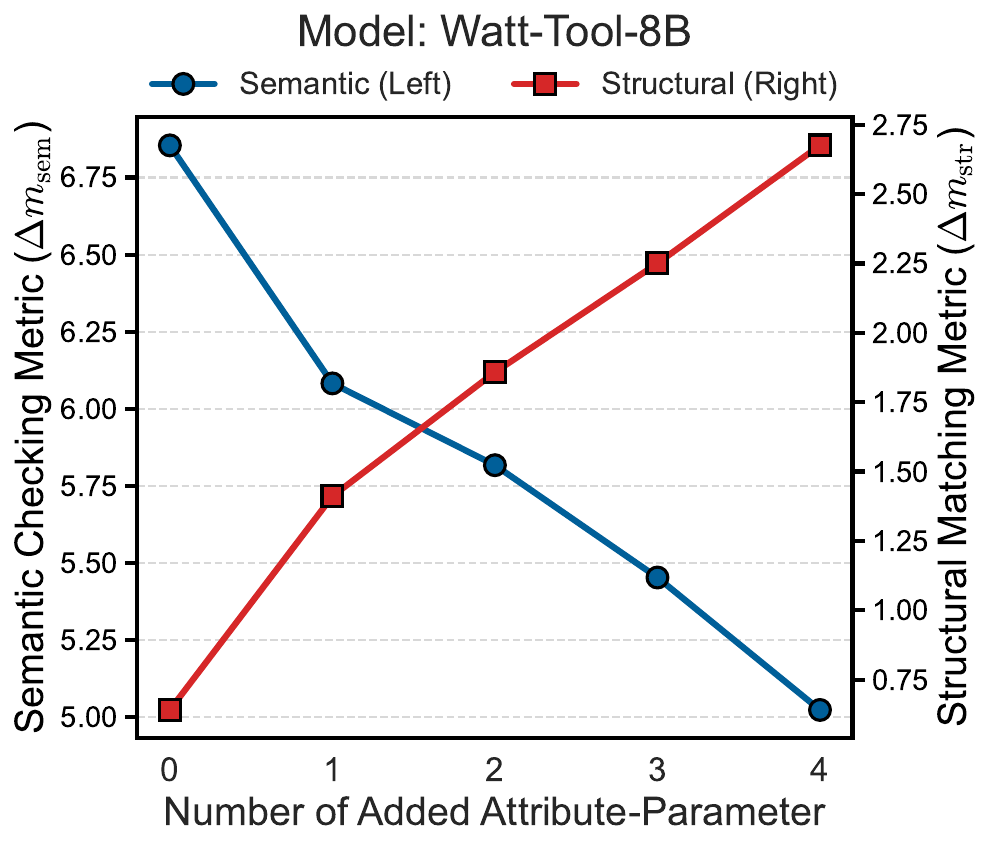}
    \caption{Pathway strengths across degrees of structural alignment for Qwen3-8B and ToolACE-2.5-8B.}
    \label{fig:dual_trend}
\end{figure*}

Here, we further investigate how semantic and structural pathways vary with increasing structural alignment.
Across $\mathcal{D}_0,\ldots,\mathcal{D}_4$, we zero-ablate the top-$k$ attention weights of the semantic and structural pathways, where $k$ equals the 2\% of the LLM attention heads, and report the results in Fig.~\ref{fig:dual_trend}. These results further indicate that the relative strength of the semantic and structural pathways drives LLMs' tool invocation decisions.

\section{Pathways Intervention}
\label{app:pathway_intervention}
\begin{figure*}[t]
    \centering
    \includegraphics[width=0.48\linewidth]{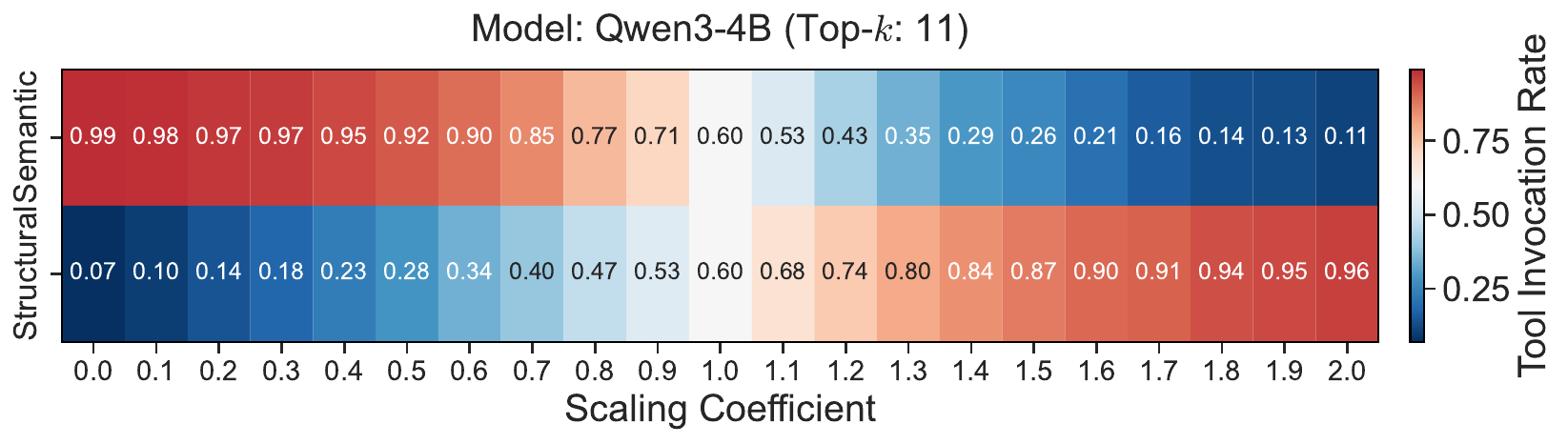}
    \includegraphics[width=0.48\linewidth]{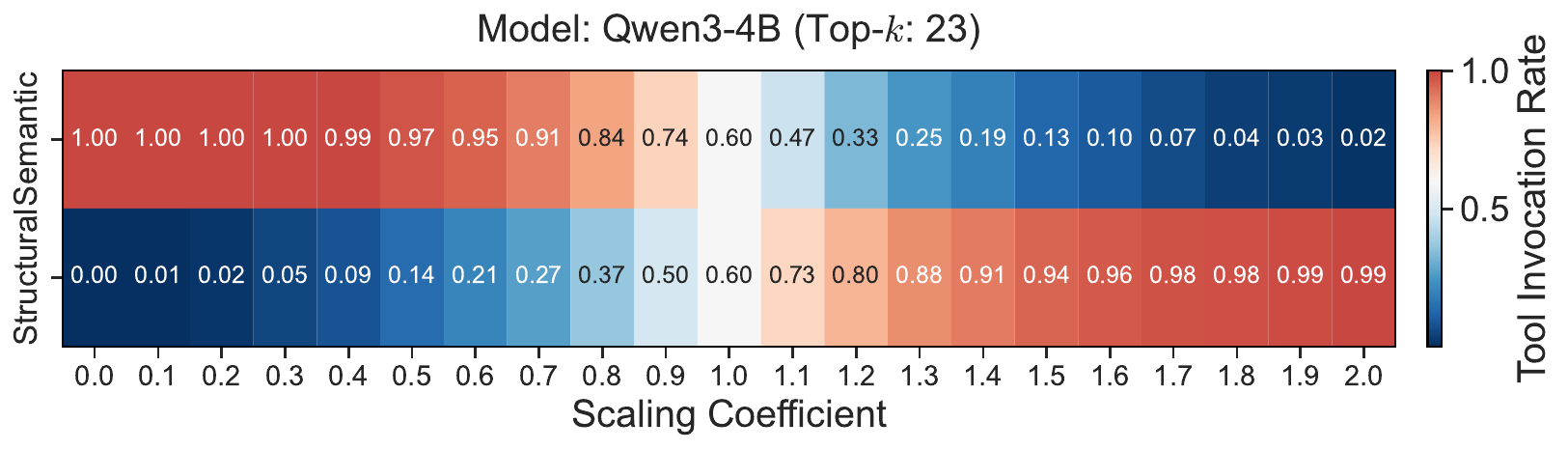}
    
    \includegraphics[width=0.48\linewidth]{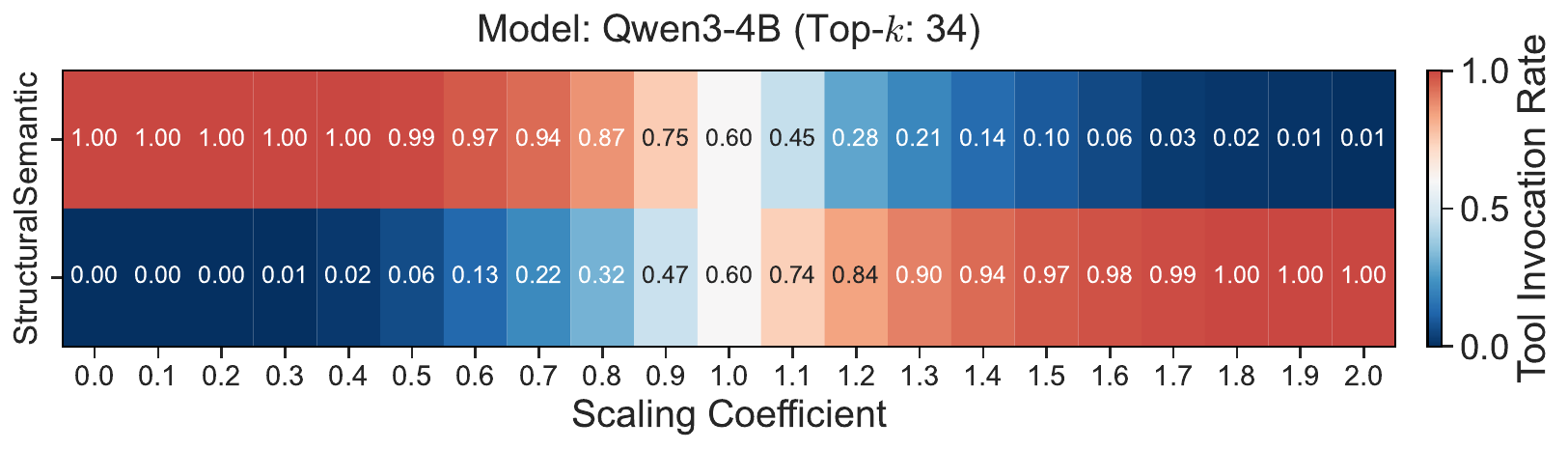}
    \caption{TIR at different top-$k$ thresholds under varying scaling coefficients for Qwen3-4B.}
    \label{fig:app_pathway_inter_qwen3_4b}
\end{figure*}

\begin{figure*}[t]
    \centering
    \includegraphics[width=0.48\linewidth]{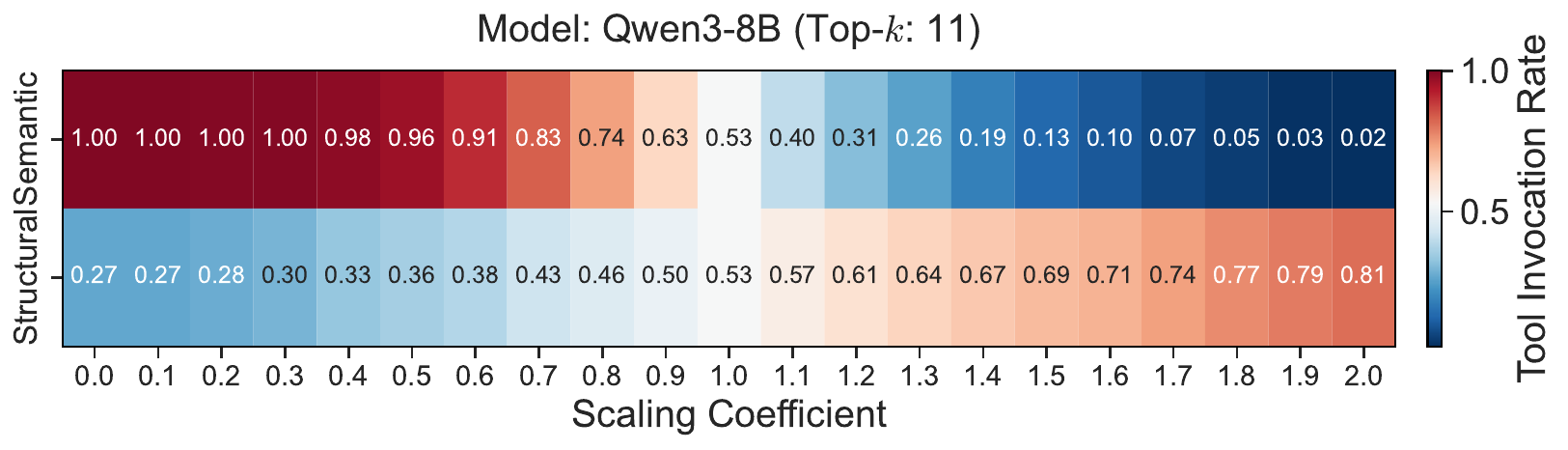}
    \includegraphics[width=0.48\linewidth]{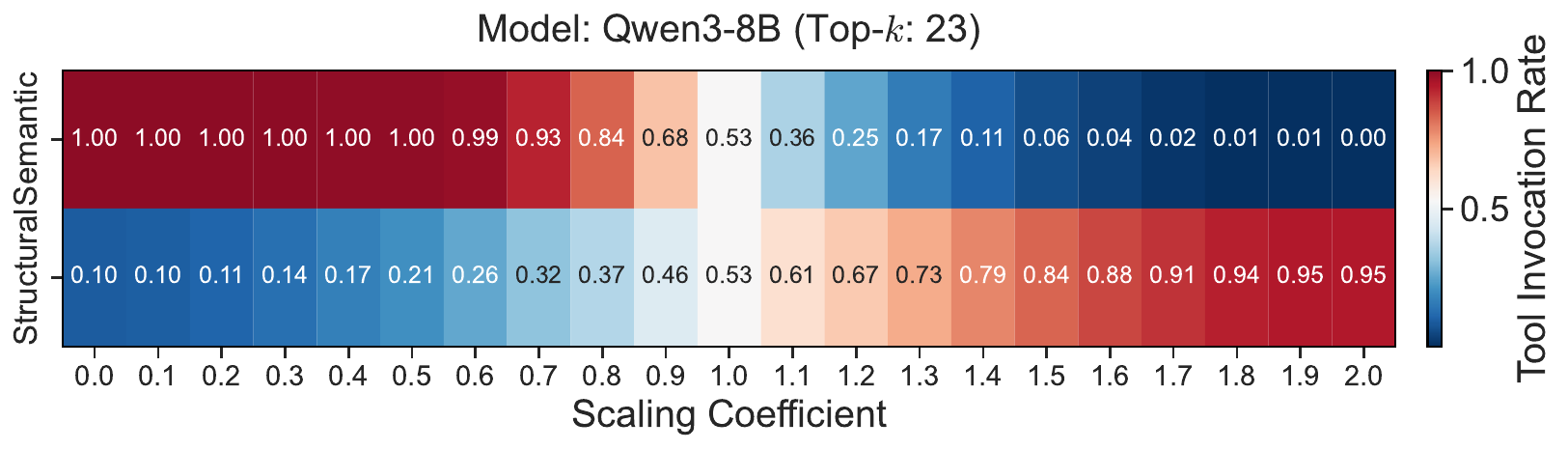}
    
    \includegraphics[width=0.48\linewidth]{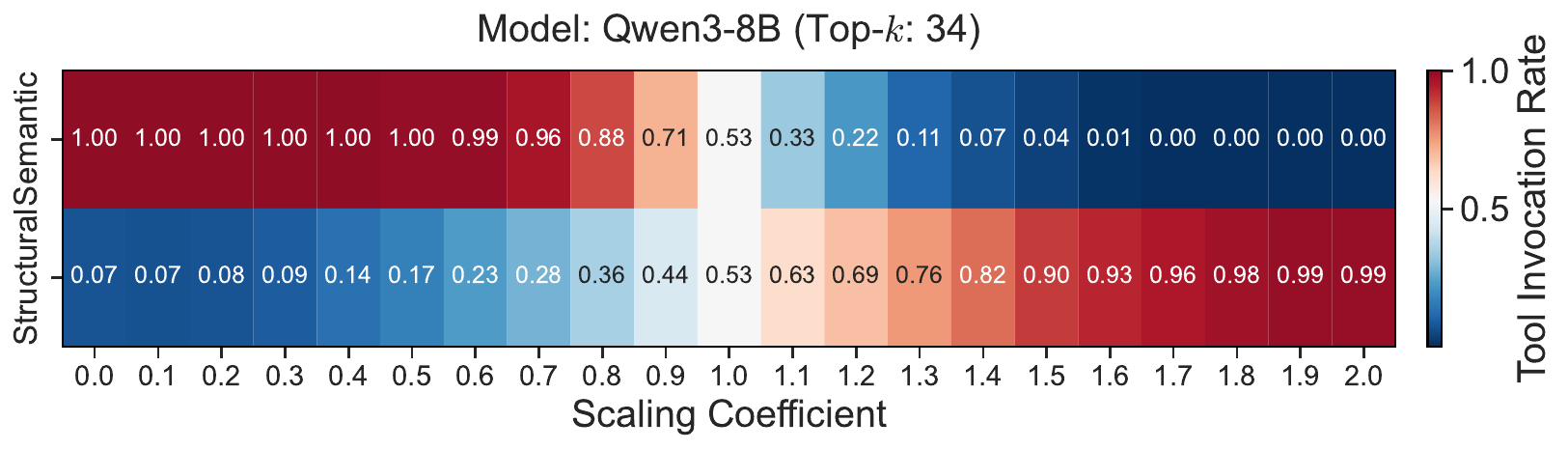}
    \caption{TIR at different top-$k$ thresholds under varying scaling coefficients for Qwen3-8B.}
    \label{fig:app_pathway_inter_qwen3_8b}
\end{figure*}

\begin{figure*}[t]
    \centering
    \includegraphics[width=0.48\linewidth]{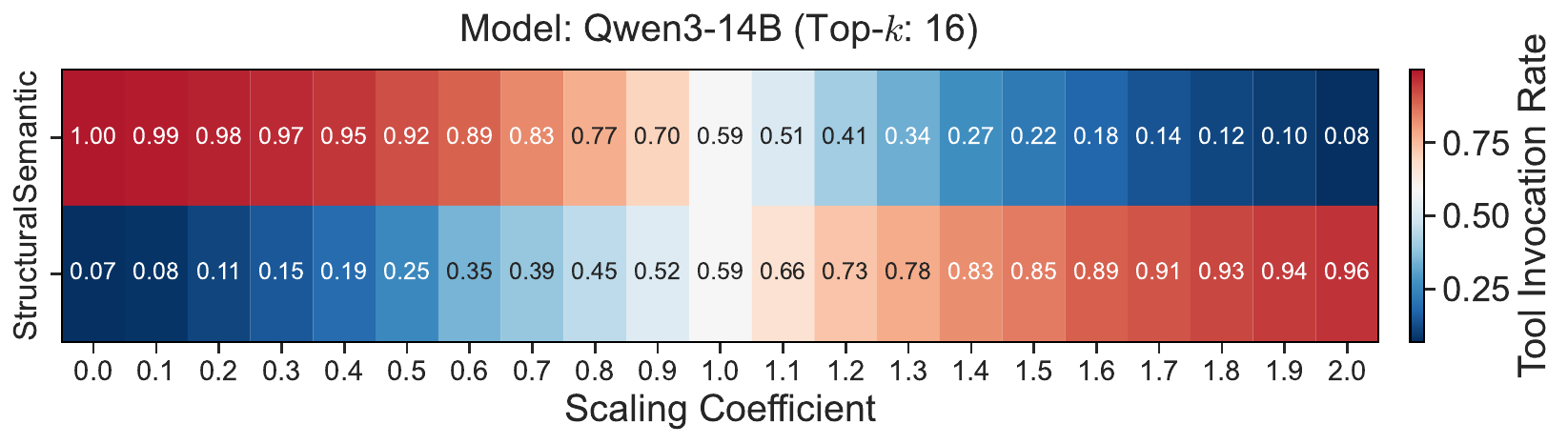}
    \includegraphics[width=0.48\linewidth]{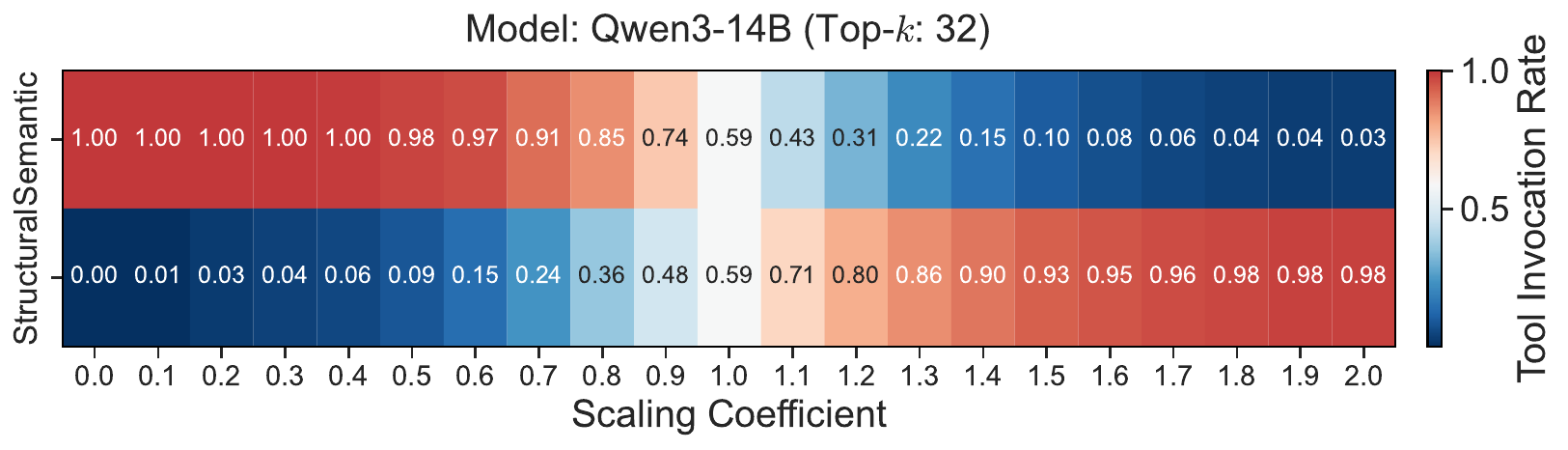}
    
    \includegraphics[width=0.48\linewidth]{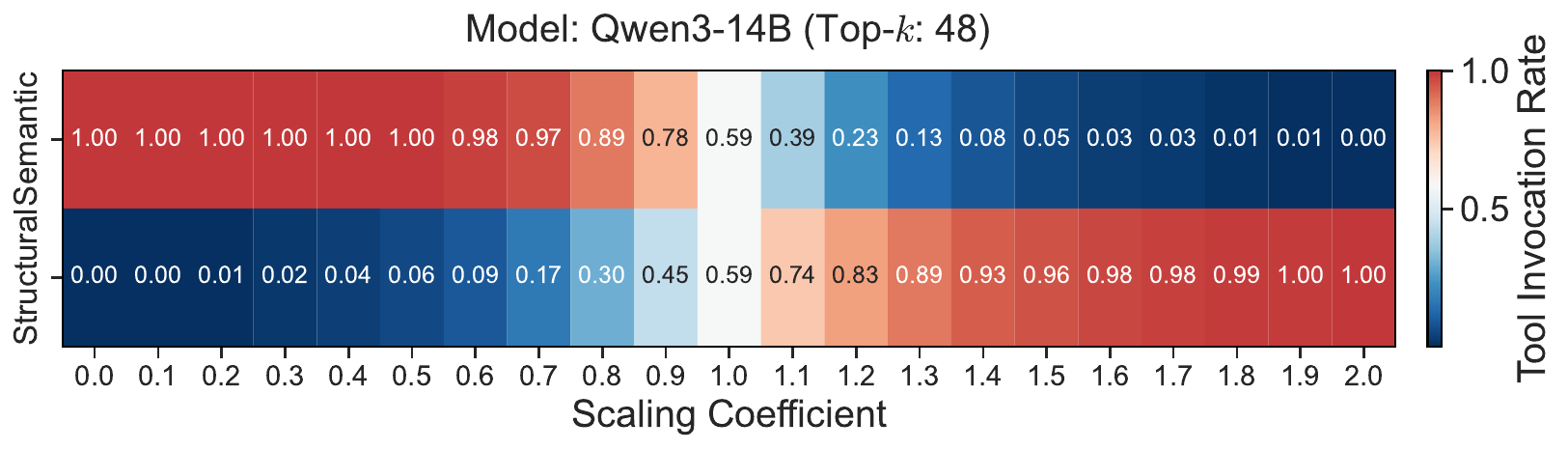}
    \caption{TIR at different top-$k$ thresholds under varying scaling coefficients for Qwen3-14B.}
    \label{fig:app_pathway_inter_qwen3_14b}
\end{figure*}

\begin{figure*}[t]
    \centering
    \includegraphics[width=0.48\linewidth]{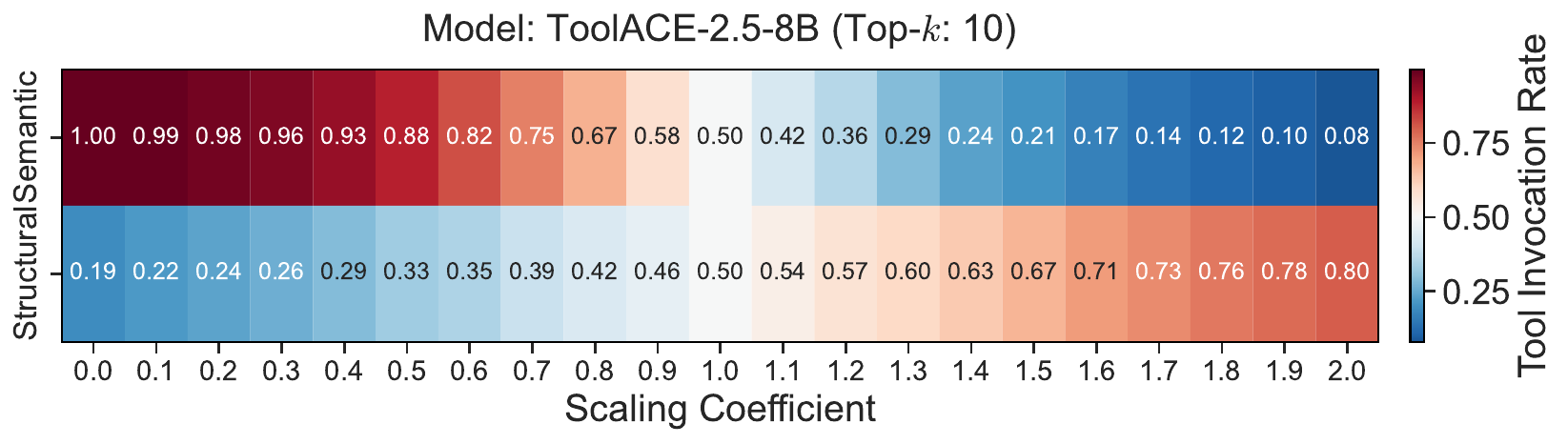}
    \includegraphics[width=0.48\linewidth]{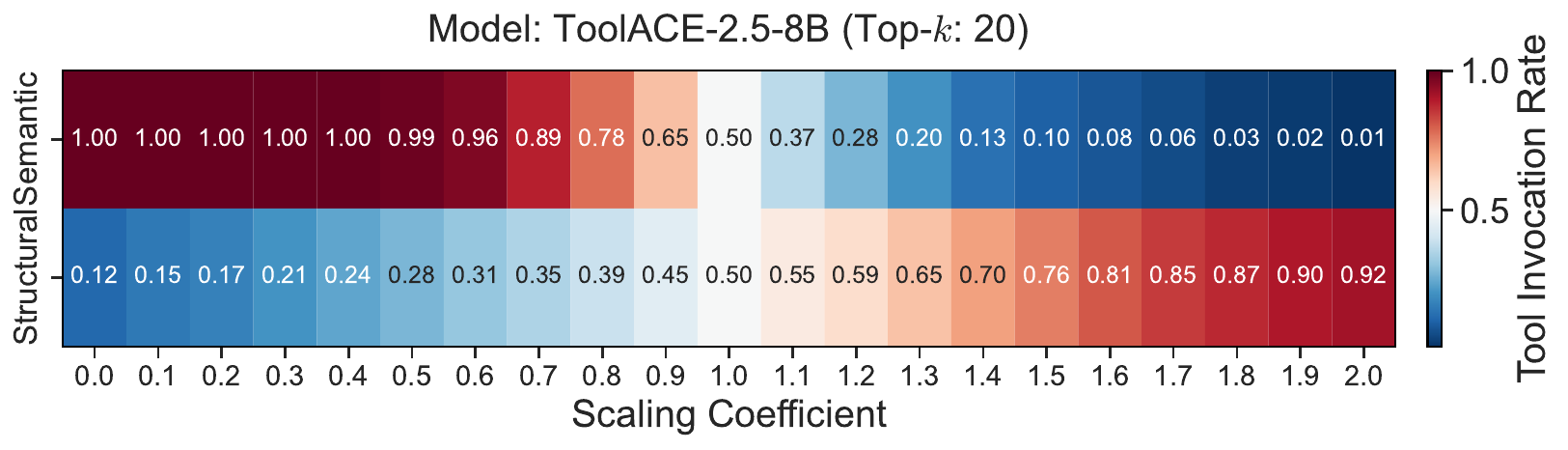}
    
    \includegraphics[width=0.48\linewidth]{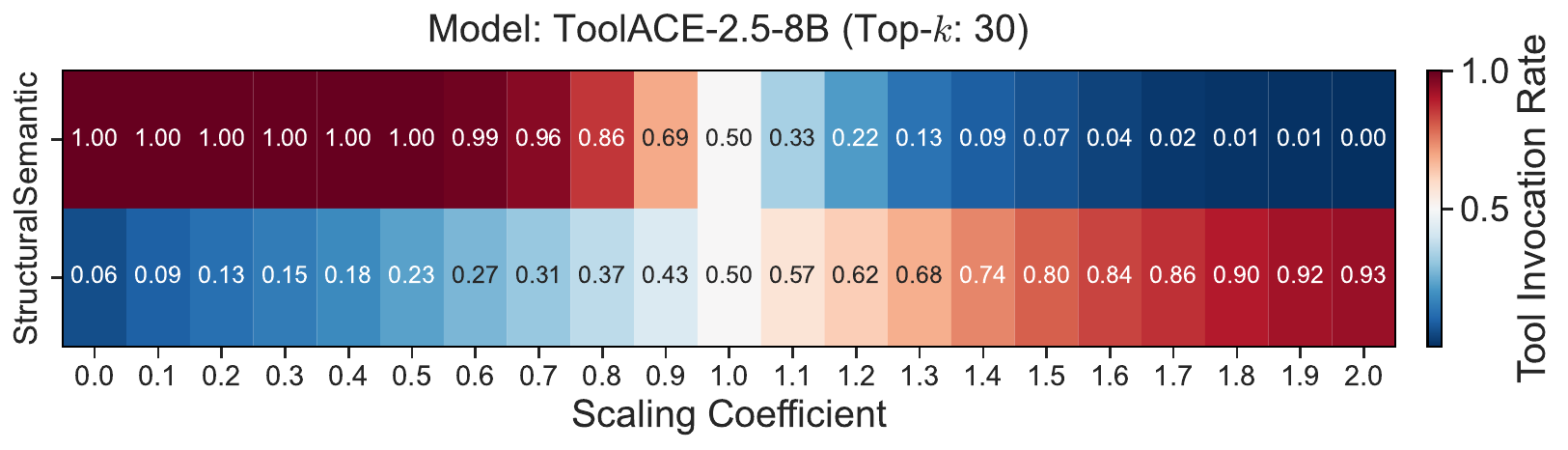}
    \caption{TIR at different top-$k$ thresholds under varying scaling coefficients for ToolACE-2.5-8B.}
    \label{fig:app_pathway_inter_toolace}
\end{figure*}

\begin{figure*}[t]
    \centering
    \includegraphics[width=0.48\linewidth]{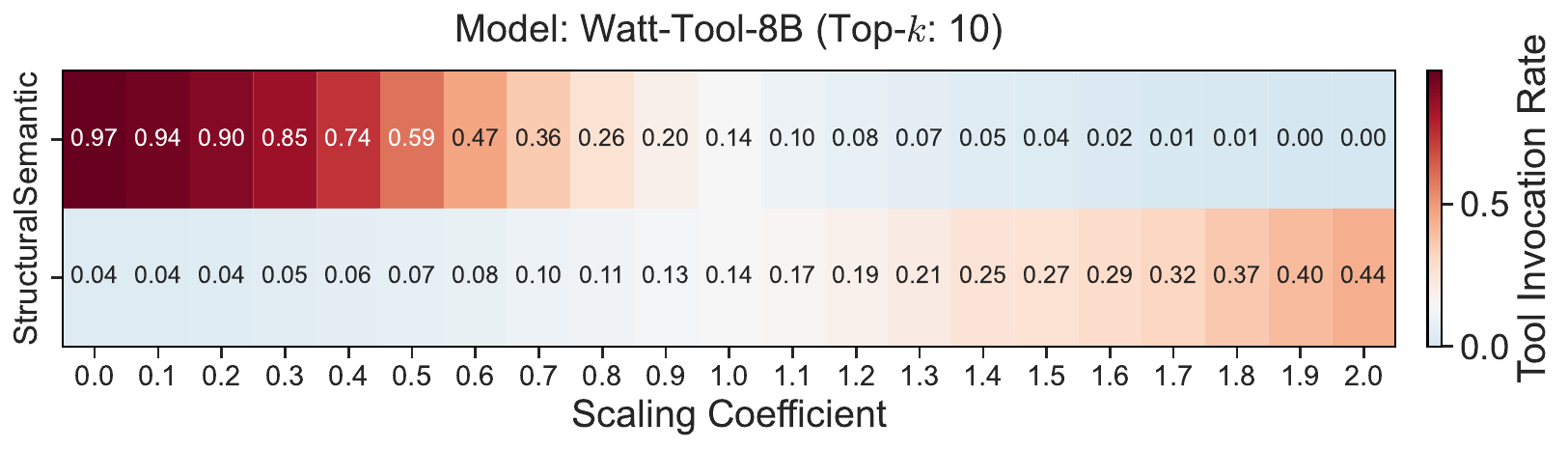}
    \includegraphics[width=0.48\linewidth]{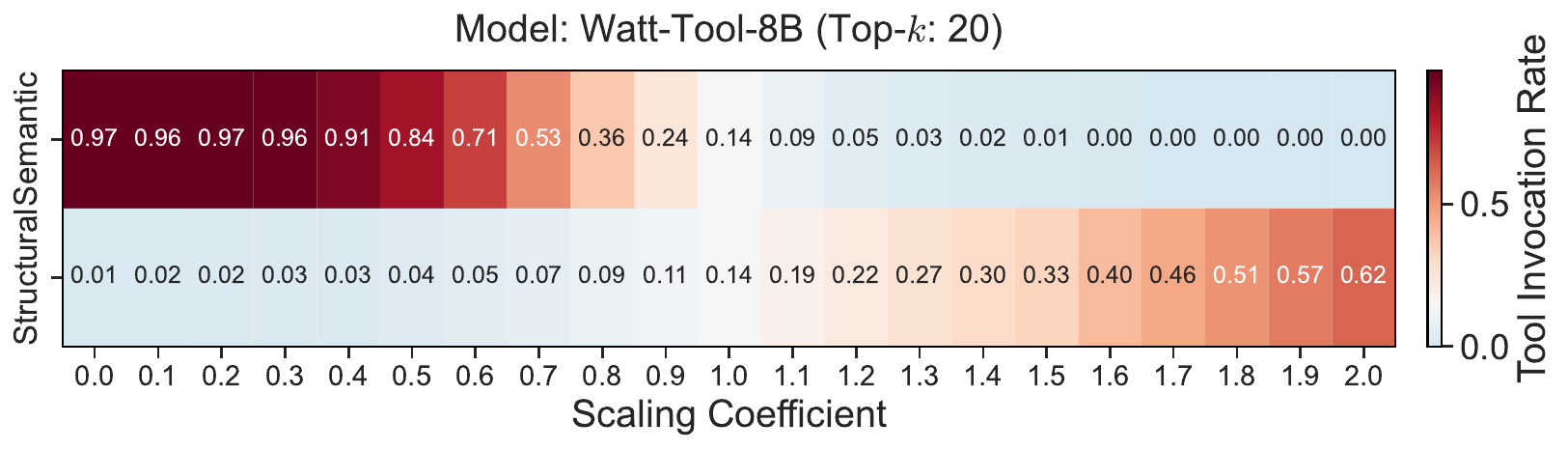}
    
    \includegraphics[width=0.48\linewidth]{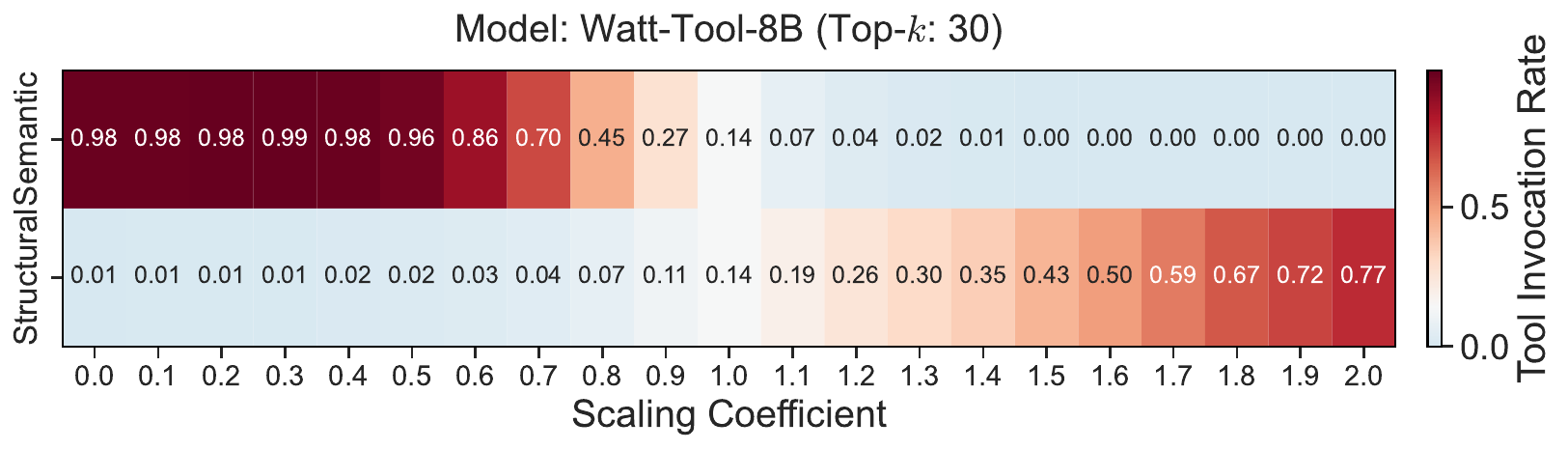}
    \caption{TIR at different top-$k$ thresholds under varying scaling coefficients for Watt-Tool-8B.}
    \label{fig:app_pathway_inter_watt}
\end{figure*}

We further examine other models and alternative top-$k$ thresholds, and report the variation in TIR under different scaling factors, as shown in Figs.~\ref{fig:app_pathway_inter_qwen3_4b},~\ref{fig:app_pathway_inter_qwen3_8b},~\ref{fig:app_pathway_inter_qwen3_14b},~\ref{fig:app_pathway_inter_toolace},~\ref{fig:app_pathway_inter_watt}.

\section{Selection of Scaling Coefficient}
\label{app:coefficient}
We select the optimal $\rho_{\text{sem}}$ and $\rho_{\text{str}}$ on the validation set by grid search: $\rho_{\text{sem}}\in\{1.1,1.2,1.3,1.4,1.5\}$ and $\rho_{\text{str}}\in\{0.5,0.6,0.7,0.8,0.9\}$. We retain combinations that achieve at least a 60\% relative reduction in TIR on the original samples, while the increase in TIR on semantic counterfactuals is no more than 3\%\footnote{We relax the 3\% threshold to 6\% since no combination is feasible for Qwen3-4B.}. Among the remaining candidates, we select the combination that minimizes the sum of TIR on the original samples and the non-invocation rate on semantic counterfactual samples.

\section{Prompt Baseline}
\label{app:prompt_baseline}
We additionally append the following instruction to the end of the system prompt: 
``\textit{If none of the tools can be used, point it out.}''
We then evaluate LLMs under this setting.

\section{Additional Analysis on Qwen3 Models}
\subsection{Impact of Thinking Mode}
\label{app:thinking_mode}
We evaluate the impact of enabling the thinking mode on Qwen3-8B and 14B models. Specifically, we measure TIR on a subset of $\mathcal{D}_1$ (randomly selecting 10 tool templates and 5 sibling pairs per template) under the ``Substitute'' intervention strategy, and compare the results against the default non-thinking setting on the same data points. As shown in Table~\ref{tab:thinking_mode}, test-time scaling via the thinking mode reduces erroneous tool invocations, particularly for the 14B model. However, directly comparing reasoning and non-reasoning modes is not entirely fair, as the former consumes significantly more compute and response time. Moreover, the error rates under the thinking mode remain substantial (30.8\% and 29.2\%), further highlighting the prevalence of structural alignment bias.

\begin{table}[h]
\centering
\small
\resizebox{\columnwidth}{!}{
\begin{tabular}{lcccc}
\toprule
\multirow{2}{*}{Strategy} & \multicolumn{2}{c}{Qwen3-8B} & \multicolumn{2}{c}{Qwen3-14B} \\
\cmidrule(lr){2-3} \cmidrule(lr){4-5}
& Disabled & Enabled & Disabled & Enabled \\
\midrule
Original & 47.2 & \textbf{30.8} & 50.8 & \textbf{29.2} \\
Substitute & 21.6 & \textbf{15.6} & 26.8 & \textbf{14.0} \\
\bottomrule
\end{tabular}
}
\caption{TIR (\%) with thinking mode disabled and enabled on Qwen3-8B and 14B models, evaluated on a subset of $\mathcal{D}_1$.}
\label{tab:thinking_mode}
\end{table}

\subsection{Impact of Model Size}
\label{app:model_size}
We further evaluate Qwen3-32B on $\mathcal{D}_1$ using the same settings as Fig.~\ref{fig:tir_cf}. The results are shown in Table~\ref{tab:model_size}. While Qwen3-8B exhibits a notably lower TIR, this is not observed in the larger Qwen3-32B, which instead yields the highest error rate (62.22\%) across all model sizes. Overall, TIR does not change monotonically with model size: mid-sized models (8B, 14B) show lower rates, whereas both the smaller (4B) and larger (32B) models exhibit higher rates.

\begin{table}[t]
\centering
\resizebox{\columnwidth}{!}{
\begin{tabular}{p{1.4cm}M{1.5cm}M{1.5cm}M{1.5cm}M{1.5cm}}
\toprule
Strategy & Qwen3-4B & Qwen3-8B & Qwen3-14B & Qwen3-32B \\
\midrule
Original & 60.44 & 53.43 & 60.40 & \textbf{62.22} \\
Add & 52.85 & 46.20 & 50.24 & \textbf{55.41} \\
Remove & 33.80 & 26.08 & 35.27 & \textbf{41.52} \\
Substitute & 32.08 & 23.17 & 29.90 & \textbf{34.47} \\
\bottomrule
\end{tabular}
}
\caption{TIR (\%) across different Qwen3 model sizes on $\mathcal{D}_1$.}
\label{tab:model_size}
\end{table}

\subsection{Additional Baselines}
\label{app:additional_baselines}

\begin{table}[t]
\centering
\resizebox{\columnwidth}{!}{
\begin{tabular}{lcccccc}
\toprule
Method & $\mathcal{D}_0$ & $\mathcal{D}_1$ & $\mathcal{D}_2$ & $\mathcal{D}_3$ & $\mathcal{D}_4$ \\
\midrule
Base & 30.39 & 51.61 & 66.29 & 75.56 & 82.44 \\
Prompt & 19.90 & 34.29 & 48.73 & 57.66 & 66.20 \\
\midrule
CAA (Ours) & \underline{4.29} & \underline{6.49} & \underline{7.27} & \underline{8.98} & \underline{11.12} \\
\midrule
One-shot (r) & 63.71 & 84.05 & 91.22 & 94.78 & 96.98 \\
One-shot (s) & 30.54 & 49.37 & 60.14 & 69.80 & 79.95 \\
Fine-tuning & \textbf{1.41} & \textbf{2.48} & \textbf{2.10} & \textbf{3.51} & \textbf{4.39} \\
\bottomrule
\end{tabular}
}
\caption{TIR (\%) of additional baselines on Qwen3-8B across SABEval. (r) and (s) denote raw and summarized one-shot formats, respectively.}
\label{tab:additional_baselines}
\end{table}

We compare our method with one-shot prompting and supervised fine-tuning on Qwen3-8B. The results are shown in Table~\ref{tab:additional_baselines}.

\paragraph{One-shot Prompting.}
We randomly select one training sample to demonstrate the correct refusal behavior, and evaluate two formats: (1)~\textit{One-shot (raw)}, which includes the complete tool definition, the user query, and the expected refusal response; and (2)~\textit{One-shot (summarized)}, which provides a natural language description of the refusal logic (i.e., ``If a tool is designed to retrieve the most visited zoo, but the user asks for the most visited museum, you should not call the tool and instead explicitly state it is unsuitable.''). Note that the ``Prompt'' method in Table~\ref{tab:main_result} serves as the zero-shot baseline. Our results show that one-shot prompting fails to improve performance. Surprisingly, the raw format significantly degrades it, suggesting that exposing the model to additional complete structural examples exacerbates structural alignment bias rather than mitigating it.

\paragraph{Supervised Fine-Tuning.}
We fine-tune Qwen3-8B using LoRA (rank=8, $\alpha$=8, dropout=0.05) on all parameters and select the best checkpoint over 2 epochs. While fine-tuning achieves strong in-distribution performance on SABEval, it fundamentally acts as a black box that forces the model to fit refusal behaviors. Our method approaches the efficacy of fine-tuning without requiring any parameter updates, and allows flexible control over intervention strength at test time. Moreover, our method is interpretable. It serves not merely as a mitigation strategy, but as a direct causal validation of the competing semantic and structural pathways. This mechanistic understanding may also lay the groundwork for more effective future mitigation strategies, such as targeted fine-tuning on the attention heads within the identified pathways.

%%%%%%%%%%%%%%%%%%%%%%%%%%%%%%%%%%

%%%%%%%%%%%%%%

%%%%%%%%%%%%%%%%%%%%%%%%%%%%%%%%%%

%%%%%%%%%%%%%%%%%%%%%%%%%%%%%%%%%%
\begin{table*}[t]
\centering
\small
\renewcommand{\tabularxcolumn}[1]{p{#1}}
\rowcolors{2}{white}{gray!10}
\begin{tabularx}{\linewidth}{l >{\raggedright\arraybackslash}X}
    \toprule
    \textbf{Span Name}  &  \textbf{Span Content} \\
    \midrule
    system-start & ``<|im\_start|>system\textbackslash n'' \\
    tool-instruction & ``\# Tools\textbackslash n\textbackslash nYou may call one or more functions to assist with the user query.\textbackslash n\textbackslash nYou are provided with function signatures within <tools></tools> XML tags:\textbackslash n'' \\
    tool-start & ``<tools>\textbackslash n'' \\
    tool-definition & \textcolor{red}{tool definition}\\
    tool-end & ``</tools>\textbackslash n\textbackslash n'' \\
    output-instruction & For each function call, return a json object with function name and arguments within <tool\_call></tool\_call> XML tags:\textbackslash n<tool\_call>\textbackslash n{"name": <function-name>, "arguments": <args-json-object>}\textbackslash n</tool\_call> \\
    system-end & ``<|im\_end|>\textbackslash n'' \\
    user-start & ``<|im\_start|>user\textbackslash n'' \\
    user-query & \textcolor{red}{user query} \\
    user-end & ``<|im\_end|>\textbackslash n'' \\
    assistant-start & ``<|im\_start|>assistant\textbackslash n'' \\
    response & ``<think>\textbackslash n\textbackslash n</think>\textbackslash n\textbackslash n'' \\
    \bottomrule
\end{tabularx}
\caption{Spans for Qwen3 models (4B, 8B, and 14B).}
\label{tab:span_qwen3}
\end{table*}

\begin{table*}[t]
\centering
\small
\renewcommand{\tabularxcolumn}[1]{p{#1}}
\rowcolors{2}{white}{gray!10}
\begin{tabularx}{\linewidth}{l >{\raggedright\arraybackslash}X}
    \toprule
    \textbf{Span Name}  &  \textbf{Span Content} \\
    \midrule
    BOS & ``<|begin\_of\_text|>'' \\
    system-start & ``<|start\_header\_id|>system<|end\_
    header\_id|>\textbackslash n\textbackslash n'' \\
    background & ``Cutting Knowledge Date: December 2023\textbackslash nToday Date: ...\textbackslash n\textbackslash n''\\ 
    % \midrule
    tool-instruction & ``You are an expert in composing functions. You are given a question and a set of possible functions. Based on the question, you will need to make one or more function/tool calls to achieve the purpose.\textbackslash nIf none of the function can be used, point it out. If the given question lacks the parameters required by the function, also point it out.\textbackslash nYou should only return the function call in tools call sections.\textbackslash n\textbackslash n'' \\
    output-instruction & ``If you decide to invoke any of the function(s), you MUST put it in the format of [func\_name1(params\_
    name1=params\_value1, params\_name2=params\_value2...), func\_name2(params)]\textbackslash nYou SHOULD NOT include any other text in the response.\textbackslash n'' \\
    tool-start & ``Here is a list of functions in JSON format that you can invoke.\textbackslash n'' \\
    tool-definition & \textcolor{red}{tool definition}\\
    system-end & ``<|eot\_id|>'' \\
    user-start & ``<|start\_header\_id|>user<|end\_header\_id|>\textbackslash n\textbackslash n'' \\
    user-query & \textcolor{red}{user query} \\
    user-end & ``<|eot\_id|>'' \\
    assistant-start & ``<|start\_header\_id|>assistant<|end\_header\_id|>\textbackslash n\textbackslash n'' \\
    \bottomrule
\end{tabularx}
\caption{Spans for ToolACE-2.5-8B and Watt-Tool-8B models (identical partitioning, except Watt-Tool-8B has no background span).}
\label{tab:span_toolace}
\end{table*}

%%%%%%%%%%%%%%%%%%%%%%%%%%%%%%
\begin{figure*}[t]
\begin{tcolorbox}[
    colframe=black!75,
    colback=gray!5,
    coltitle=white,
    fonttitle=\bfseries,
    title={Annotation Example: Base Class, Derived Class, Tool Template}
]
\begin{lstlisting}
{
    @\textcolor{blue}{"base\_class":}@ "athlete",
    @\textcolor{blue}{"derived\_class":}@ [
        "basketball",
        "soccer",
        "baseball",
        "american football",
        "ice hockey",
        "volleyball",
        "esports",
        "cricket",
        "rugby union",
        "handball",
        "futsal",
        "field hockey"
    ],
    @\textcolor{blue}{"tool\_template":}@ {
        "name": "find_<class>_athlete",
        "description": "Find the profile information of a <class> 
                        athlete based on their full name.",
        "parameters": {
            "type": "object",
            "properties": {
                "name": {
                    "type": "string",
                    "description": "The full name of the <class> athlete."
                }
            },
            "required": [
              "name"
            ]
        }
    }
}

\end{lstlisting}
\end{tcolorbox}
\caption {An example annotation illustrating a base class, derived classes, and a tool template.}
\label{fig:dataset_tool_example}
\end{figure*}

\begin{figure*}[t]
\begin{tcolorbox}[
    colframe=black!75,
    colback=gray!5,
    coltitle=white,
    fonttitle=\bfseries,
    title={Prompt for Query Generation During Dataset Construction}
]
\small
\begin{lstlisting}
Your task is to generate a set of diverse user queries for the given tool.
Note that the queries you generate represent real user queries directed at an LLM assistant. The LLM assistant will need to call the provided tool to answer these queries.

Rules for Query Generation:
1. Derived Class Inclusion: Every generated query must contain the exact derived class name provided.
2. Parameter Constraints: 
  - Each query must explicitly contain information for all and only the parameters of the
    tool.
  - The expected parameter values must be specific and realistic. Avoid vague values.
3. No Attachments: Do not assume or pretend that files, images, audio clips, videos, or any other attachments are being provided.
4. Quality: Generated queries must be solvable with the tool without requiring further clarification.
5. Diversity:
  - You should generate at least 5 distinct queries.
  - The queries should have varied sentence structures (e.g., imperative commands,
    interrogative queries).
  - The parameter values across different queries should also be diverse, covering a wide
    range of realistic scenarios, if applicable.

Output Format:
Return a single JSON array as follows:
[ {{ "query": "First generated query..." }}, {{ "query": "Second generated query..." }}, ... ]

[EXAMPLE START]
@\textcolor{red}{\{few-Shot examples\}}@
[EXAMPLE END]

Now, generate queries for the following tool.

Tool Definition:
@\textcolor{red}{\{tool\}}@

Derived Class Name:
@\textcolor{red}{\{subclass\}}@
\end{lstlisting}
\end{tcolorbox}
\caption {Query-generation prompt used during dataset construction.}
\label{fig:dataset_query_generation}
\end{figure*}

\begin{figure*}[t]
\begin{tcolorbox}[
    colframe=black!75,
    colback=gray!5,
    coltitle=white,
    fonttitle=\bfseries,
    title={Prompt for Adding Additional Parameters }
]
\small
\begin{lstlisting}
Your task is to propose a set of new, additional parameters for the given tool template.
These parameters should enrich the tools's functionality while remaining applicable across all
derived classes.

You are Given:
1. Tool Template: An tool schema that uses '<class>' as a placeholder for a specific derived class.
2. List of Derived Classes: A list of the specific derived class names that will eventually replace '<class>'.
3. Base Class Description: A brief explanation of the base class corresponding to the tool template.

Design Principles:
- The tool's purpose must remain clear and unambiguous after adding new parameters.
- All parameter values would be provided by the user when invoke the tool, not generated or
  assumed by the LLM assistant. The LLM acts as a bridge to execute the tool with user-
  provided information, not as a param value generator.

Rules:
1. Universally Applicable: Each proposed parameter must be universally applicable and make sense for all derived classes provided.
2. Uniqueness: The proposed parameters must be entirely new. They cannot duplicate the functionality or name of any parameters already present in the tool template.
3. Placeholder Usage: Parameter names must be generic and must not contain the '<class>' placeholder. Parameter description, however, should contain the '<class>' placeholder if it is
contextually appropriate when replaced with a specific subclass.
4. Quantity: Generate at least four distinct and meaningful parameters.
5. Type: New parameters should be simple types: string, integer, number, boolean or array of simple types. Do not propose complex nested structures.

Output Format:
Return a single JSON object as follows:
{{"parameter_name_1": {{"type": "string", "description": "Description for param 1."}}, "parameter_name_2": {{"type": "integer", "description": "Description for param 2."}}, ...}}

[EXAMPLE START]
@\textcolor{red}{\{few-Shot examples\}}@
[EXAMPLE END]

Now, generate new parameters based on the following inputs.

Tool Template:
@\textcolor{red}{\{tool template\}}@

List of Derived Classes:
@\textcolor{red}{\{derived class list\}}@

Base Class Description:
@\textcolor{red}{\{base class description\}}@
\end{lstlisting}
\end{tcolorbox}
\caption {Prompt for adding additional parameters during dataset extension.}
\label{fig:dataset_extension}
\end{figure*}

%%%%%%%%%%%%%%%%%%%%%%%%%%%%%%%%%%%%%%%%%%%%% Qwen3
\begin{figure*}[t]
\begin{tcolorbox}[
    colframe=black!75,
    colback=gray!5,
    coltitle=white,
    fonttitle=\bfseries,
    title={System Prompt (Qwen3-4B, Qwen3-8B and Qwen3-14B)}
]
\begin{lstlisting}
# Tools

You may call one or more functions to assist with the user query.

You are provided with function signatures within <tools></tools> XML tags:
<tools>
@\textcolor{red}{\{tool\_definition\}}@
</tools>

For each function call, return a json object with function name and arguments
within <tool_call></tool_call> XML tags:
<tool_call>
{"name": <function-name>, "arguments": <args-json-object>}
</tool_call>
\end{lstlisting}
\end{tcolorbox}
\caption {Default system prompt for Qwen3-4B, Qwen3-8B and Qwen3-14B}
\label{fig:qwen_prompt}
\end{figure*}

%%%%%%%%%%%%%%%%%%%%%%%%%%%%%%%%%%%%%%%%%%%%% ToolACE-8B and Watt-Tool-8B
\begin{figure*}[t]
\begin{tcolorbox}[
    colframe=black!75,
    colback=gray!5,
    coltitle=white,
    fonttitle=\bfseries,
    title={System Prompt (ToolACE-8B and Watt-Tool-8B)}
]
\begin{lstlisting}
You are an expert in composing functions. You are given a question and a set of possible functions. Based on the question, you will need to make one or more function/tool calls to achieve the purpose.
If none of the function can be used, point it out. If the given question lacks the parameters required by the function, also point it out.
You should only return the function call in tools call sections.

If you decide to invoke any of the function(s), you MUST put it in the format
of [func_name1(params_name1=params_value1, params_name2=params_value2...), func_name2(params)]
You SHOULD NOT include any other text in the response.
Here is a list of functions in JSON format that you can invoke.
@\textcolor{red}{\{tool\_definition\}}@
\end{lstlisting}
\end{tcolorbox}
\caption {Default system prompt for ToolACE-8B and Watt-Tool-8B}
\label{fig:toolace_prompt}
\end{figure*}

\end{document}